\documentclass{article}
\usepackage{todonotes}
\usepackage{authblk}
\usepackage{graphicx}
\usepackage{ulem}
   
\usepackage{amssymb}
\usepackage{lineno}
\usepackage[margin=0.7in]{geometry}
\usepackage{caption}
\graphicspath{{fig/}}
\usepackage{graphicx}
\usepackage{booktabs}
\usepackage[flushleft]{threeparttable}
\usepackage{graphicx}
\usepackage{setspace}
\usepackage{multirow}
\usepackage{subcaption}
\usepackage{bigstrut}
\usepackage[title]{appendix}
\usepackage{verbatim}
\usepackage{array,tabularx,booktabs}
\usepackage{makecell}
\usepackage{boldline}
\usepackage[figuresright]{rotating}
\usepackage{adjustbox}
\usepackage{lscape}
\usepackage{amssymb,amsfonts,amsbsy,epsfig}
\usepackage{algorithm}
\usepackage[noend]{algpseudocode}
\usepackage{float}
\usepackage{rotating}
\usepackage{longtable}
\usepackage[T1]{fontenc}
\usepackage{amsmath}
\usepackage[colorlinks=true, allcolors=blue]{hyperref}
\usepackage{algorithmicx}
\usepackage{algorithm}
\usepackage{verbatim}

\title{Medical Image Classification Using Transfer Learning and Chaos Game Optimization on the Internet of Medical Things}

\date{}

\author[1]{\small Alhassan Mabrouk}
\author[2]{\small Abdelghani Dahou}

\author[3,4,5]{\small Mohamed Abd Elaziz}
\author[6]{\small Rebeca P. Díaz Redondo}
\author[7]{\small Mohammed Kayed}

\affil[1]{\footnotesize Mathematics and Computer Science Department, Faculty of Science, Beni-Suef University, Beni Suef 62511, Egypt}
\affil[2]{\footnotesize Mathematics and Computer Science department, University of Ahmed DRAIA, 01000, Adrar, Algeria}
\affil[3]{\footnotesize Department of Mathematics, Faculty of Science, Zagazig University, Zagazig, 44519, Egypt}
\affil[4]{\footnotesize Faculty of Computer Science and Engineering, Galala University, Suez 435611, Egypt}
\affil[5]{\footnotesize Artificial Intelligence Research Center (AIRC), Ajman University, Ajman P.O. Box 346, United Arab Emirates}

\affil[6]{\footnotesize Information \& Computing Lab, AtlanTTIC Research Center, Telecommunication Engineering School, Universidade de Vigo, 36310 Vigo, Spain}
\affil[7]{\footnotesize Computer Science Department, Faculty of Computers and Artificial Intelligence, Beni-Suef University, Beni Suef 62511, Egypt}

\providecommand{\keywords}[1]
{
  \small	
  \textbf{\textit{Keywords---}} #1
}

\begin{document}

\maketitle

\begin{abstract}

The Internet of Medical Things (IoMT) has dramatically benefited medical professionals that patients and physicians can access from all regions.
\color{black}
Although the automatic detection and prediction of diseases such as melanoma and leukemia is still being researched and studied in IoMT, existing approaches are not able to achieve a high degree of efficiency. Thus, with a new approach that provides better results, patients would access the adequate treatments earlier and the death rate would be reduced.
\color{black}
\color{black}
Therefore, this paper introduces an IoMT proposal for medical images classification that may be used anywhere, i.e. it is an ubiquitous approach. It was design in two stages: first, we employ a Transfer Learning (TL)-based method for feature extraction, which is carried out using MobileNetV3; second, we use the Chaos Game Optimization  (CGO) for feature selection, with the aim of excluding unnecessary features and improving the performance, which is key in IoMT.
\color{black}
Our methodology was evaluated using ISIC-2016, PH2, and Blood-Cell datasets. The experimental results indicated that the proposed approach obtained an accuracy of 88.39\% on ISIC-2016, 97.52\% on PH2, and 88.79\% on Blood-cell. Moreover, our approach had successful performances for the metrics employed compared to other existing methods.

\end{abstract}

\keywords{Internet of Things; Image Processing; Deep learning; Skin Cancer; Feature Selection Optimization Algorithms.}

\section{Introduction}
\color{black}
The Internet of Things (IoT) has been formulated to define the use of devices that can be controlled remotely \cite{ren2022application}. The development of these devices allowed a wide range of uses. \color{black}Hence, IoT is used in many areas, such as industrial \cite{alharbi2021botnet}, smart cites \cite{mehra2022lbecr}, agriculture \cite{sinha2022recent}, and Internet of Medical Things (IoMT) \cite{ tamulis2022affective}\color{black}. However, IoMT technology has been commonly applied due to its high performance, saving time, and efforts of specialists/patients \cite{abd2022medical}. Besides, it provides patient care, such as monitoring their medications and tracking their hospital admission location. \color{black}IoMT technologies are widely available, especially for diseases with the highest mortality rate globally, such as melanoma \cite{hossen2022federated}, leukemia \cite{karar2022intelligent}, and others\color{black}.
Technology such as mobile devices and wearables can collect information about human health to provide effective hospital care. These technologies could be used in many applications or services, like obtaining data and analyzing it, and monitoring the diagnosis of neurological illnesses. As a result of its efficiency and usability, the IoMT technology has been broadly accepted and widely used. 

\color{black}
Deep learning (DL) models can help diagnose breast cancer \cite{maqsood2022ttcnn}, pneumonia detection \cite{mabrouk2022pneumonia}, and Alzheimer's disease \cite{odusami2022intelligent} using advanced biomedical imaging methods such as thermal imaging and magnetic resonance imaging (MRI), however, these methods are expensive, require specialized medical imaging equipment, and are not available in many rural areas of developing countries. Thus, DL has recently been used by IoMT to automate and accurately diagnose a variety of diseases, that helping to facilitate efficient and appropriate healthcare \cite{ogundokun2022medical}\color{black}. 
For instance, 
an IoMT system for stroke detection using Convolution Neural Networks (CNN) and transfer learning was demonstrated to distinguish between a healthy brain and hemorrhagic and ischemic strokes in CT scan images, as introduced in \cite{taryudi2021smart}. Although, DL models outperformed traditional machine learning \cite{budd2021survey}, there is less work known for DL-based IoMT on healthcare than services available on IoMT devices. 
\color{black}
The IoMT system for stroke patients prevention can capture and maintain the patient's heartbeats, core temperature, and external factors quickly and with the required precision. 
These factors are essential for diagnosing stroke examination.
DL techniques can help prevent frequent difficulties that take much time to solve. For example, web scraping \cite{mabrouk2021seopinion}, data mining \cite{parsons2021automatic}, sentiment analysis \cite{mabrouk2020deep} are all areas where TL technology has a broad array of applications. 
\color{black}
Moreover,  these approaches need a huge size of well-labeled training data samples. Many Transfer Learning (TL)-based approaches have been developed in medical image analysis to solve this issue. Due to its capacity to effectively solve the shortcomings of reinforcement learning and supervised learning, TL is becoming more widespread in medical image processing \cite{zoetmulder2022domain}. 

TL aims to train the forecast function in the target domain by utilizing information obtained in the source domain from a vast number of labeled data sets (e.g., ImageNet). TL is widely recognized in different computer vision domains for helping to enhance the learning of sparsely labeled or limited datasets in the particular domain \cite{nguyen2022tatl}. Unfortunately, the input image properties of the training examples (i.e., a massive dataset of natural images) and the test data are highly different for TL in medical imaging (i.e., a small dataset of clinical images). Because of the significantly different domains with various and unconnected classes, as in  \cite{liu2022margin}, the transferred functions learned from the source database (training set) may be biassed when directly implemented into the target database (test set). Consequently, the biassed function's features are unlikely to be desired in the target domain, the medical image field. Moreover
TL is vital to have both indicate environmental and discriminative capability in the feature extraction process in order to improve classification accuracy \cite{ren2022robustness}. 
According to the traditional view, the TL is pre-trained in the experiment and then fine-tuned for implementation using detailed information. Unsupervised, inductive, transductive, and negative learning are all types of TL. Also, it can solve these challenges \cite{niu2021distant}. 
\color{black}
\color{black}
Hence, we use a TL model to obtain features from medical images.
\color{black}

Many features, such as color, texture, and size, are used in standard medical image categorization methods. When controlling high-dimensional feature vectors through an optimizer algorithm, the selection of optimal features is offered in a way to improve classification efficiency \cite{adel2022improving}. The optimal representation of the specified subset of features creates additional issues for the researchers. \color{black}In order to automate this method, a Feature Selection (FS) approaches have also been crucial for accurately defining these essential features.
\color{black}

Therefore, we developed a method to solve the diagnostic imaging identification challenge and optimize the process, which is wrapped as an IoMT system to reduce morbidity and mortality worldwide. \color{black} 
To the best of our knowledge, our approach is the first that tries to improve the efficiency of medical image classification on IoMT based on merging the deep learning (as MobileNetV3) and chaos game optimization metaheurstic optimization. 
\color{black}

In order to improve the performance for classifying medical images, 
the system incorporates both TL and FS optimization techniques. It is initially recommended that a TL architecture analyze the supplied medical images and develop contextualized representations without personal communication. A fine-tuned MobileNetV3 is utilized to retrieve the embedded images. Next, a novel FS method is also planned to analyze each pixel embedding and choose only the most important properties to improve medical image classification performance. The FS method depends on a new metaheuristic strategy known as Chaos Game Optimization (CGO). 
\color{black}
The reasons for employing CGO approaches to optimize the FS challenge in this paper are as follows. We would want to examine the most recent CGO optimizer. Furthermore, a CGO method is compared to the approach to complex, modern, and high-efficiency algorithms reveal that the CGO optimizer has the optimal solution for the problems examined, with typically more incredible classification performance (i.e., fewer iterations and execution time).
The contributions of this paper can be summarized as:

\begin{itemize} 

\item The proposed IoMT system helps minimize human intervention in medical centers and provides fast diagnosis reports embedded in low-resourced systems.
\item The transfer learning (i.e., MobileNetV3) model is fine-tuned on the assessed medical image datasets to extract relevant features.
\item A novel feature selection approach to select appropriate features is used to build an IoMT system. 
\item An extensive evaluation of the proposed system is reported and compared to several state-of-the-art techniques using two real-world datasets.

\end{itemize}

\color{black}

According to the paper's structure, Section \ref{rw} describes a review of recent work on medical imaging. Section \ref{pm} offers a detailed description of our approach. Section \ref{ex} analyzes the implementation results of image classification techniques. Finally, the concluding remarks give future scope in Section \ref{c}.

\section{Related works} \label{rw}

The essential strength of the classification task to help diagnose the medical image makes it an important area of research. Therefore, this section is presented the recent works about medical image classification.
\color{black}
Recently, the researchers have improved the Internet of Medical Things (IoMT) using DL and the classification task performance by applying transfer learning. Due to advances in connectivity among systems, the Internet of Things (IoT) is currently being used in various fields. When used in the medical area, the IoT can construct care and monitoring systems that could be monitored remotely. It is now possible for medical professionals and sometimes even patients to remotely access sensor data generated by devices attached to persons who are being monitored or have specific requirements \cite{manimurugan2022two}. Computed Aided Diagnosis (CAD) technologies can benefit from the IoT by providing an interaction that directly correlates the terminal to the devices for medical images classification. To put it another way, any person may now control a technology that formerly required training \cite{khelili2022iomt}. 

DL has been increasingly popular on the Internet of Medical Things (IoMT) in recent years \cite{khalil2022efficient}. As a result, the IoMT concept is suitable for building embedded technologies that can accurately diagnose disease in the same manner that professionals perform. IoMT innovation, according to \cite{kumar2022depress}, has contributed to the establishment of vital healthcare systems. Physicians may now receive it in various settings, allowing them to better diagnose patients without affecting subjective features. Another obstacle that has yet to be addressed is the disparity between rare and common diseases regarding the amount of data collected. They introduced a method to recognition of CT scan images of pulmonary and ischemic stroke on the IoMT \cite{han2020internet}. These researchers employed an IoT device to directly contact users to choose the optimum extraction methods and classifications for a given situation. 
However, it was a result of this problem that the system was underperforming. A considerable percentage of accuracy is required in the medical sector when diagnosing the form of the disease. It has been shown in previous research that early identification of cancer is vital for sick people to receive the best treatment possible. Thus, our goal is to improve medical image diagnosis by increasing the accuracy of the applied algorithms.

In recent decades, meta-heuristic optimization algorithms combined with Convolution Neural Networks (CNN) for medical image classification.
The transfer learning process has been viral, primarily since it enables the system to be more powerful, reduces financial costs, and requires fewer inputs, supported by the entry weights supplied by the training process transferred. The study \cite{cheplygina2019not} examined training from many cases through a transformation in medical image processing. The researchers discussed various types of learning and future studies possibilities. 
For fine-tuning of transfer learning, Ayan and Ünver \cite{ayan2019diagnosis} employed the Xception and VGG16 structures. When they added two fully connected layers and the multiple output tier with a SoftMax activation function, they also completely modified the architecture of Xception. In the VGG16 structure, the past eight tiers were halted, and the completely connected levels were modified. Accordingly, the testing time for each image for the VGG16 and Xception networks was 16 and 20 ms, respectively. InceptionV3, ResNet18 and GoogLeNet were among the models employed in \cite{chouhan2020novel}. Based on Convolutional networks, a determination has been made. They used each one of the models to test the premise that voting may be used to arrive at a diagnostic. In their study, the findings of the classifiers were combined using the clear majority. Accordingly, the diagnostic correlates to the class with the largest rate of start voting in the polls. The model's mean testing time per image was 161 ms using this method.
On top of that, they attained high classification rates for X-ray pictures. According to this study, pneumonia can be diagnosed using deep convolutional networks. As part of our method, we rely on classical classifiers to minimize the computing cost of classifying information.


As a result of their extensive feature representation skills, CNNs have been commonly applied in medical image processing in latest years and have shown substantial gains. Zhang et al. \cite{zhang2019medical} has developed a system for target class lesion identification based on multi-CNN collaboration. In addition, their approach was more reliable in identifying lesions, and its utility had been evaluated using necessary details. A strong ensemble structure for cancer detection was created \cite{pathan2019automated} using dynamic classification techniques. Therefore, a more distinctive and robust model can be created. To identify skin lesions on their own, in \cite{yu2020convolutional}, they proposed that a crossnet-based mixture of multiple convolutional networks may be used. For the categorization of melanomas, MobileNet and DenseNet were coupled \cite{wei2020automatic}. Because the light medical image classification model was designed to boost feature selectivity, computation complexity, and parameter settings, it differed from older systems. It used a categorization strategy that worked well.

Currently, meta-heuristic optimization algorithms are being used to solve a wide range of complex optimization problems. Rather than a single answer, a list of possible solutions allows them to navigate the solution space efficiently. They beat other optimization approaches as a result.
Ravi K Samala et al. \cite{samala2018evolutionary} suggested a method of multilayered pathway development to identify breast cancer. They used a two-stage method: transfer learning and identifying features, respectively. Region of interest (ROI) from large lesions were being used to train pre-trained CNNs. On top of it, a random forest classification model was created using the learned CNN. We evolved pathways using a Genetic Algorithm (GA) with random selection and total number crossover operators. Their research found a 34 \% change in features and a 95 \% reduction in parameter actions using their proposed strategy. Through Particle Swarm Optimization (PSO), Silva et al. \cite{da2018convolutional} optimized hyper-parameter of CNN for a false-positive reduction in CT lung images due to their comparable structures and low density, which causes false-positive results. Scientists have found that optimizing an automatic detection system can improve outcomes and minimize human intervention. In order to acquire the binary threshold value, Surbhi et al. \cite{vijh2020brain} adopted OTSU-based adaptable PSO for automatic classification of brain cancers. To reduce noise and improve the image quality, remove noise processing, and apply skull stripping. For feature extraction, GLCM was utilized, and 98 \% of the features were extracted.

Utilizing the Grey Wolf Optimization (GWO) method, Shankar K. et al. \cite{shankar2019alzheimer} developed a novel concept for Alzheimer's disease using brain imaging analysis. An initial consideration for picture editing is to remove undesirable regions. The retrieved images are then sent to CNN for feature extraction, resulting in improved performance. According to Goel et al. \cite{goel2021optconet}, OptCoNet is an optimized CNN architecture for recognizing COVID19 patients as normal/pneumonia sufferers. For hyperparameter adjustment of the convolution layer, they employed the GWO. Their study found that the proposed approach assisted in the automated examination of patients and reduced medical systems' burdens on the system. In order to enhance architectures for denoising images, Mohamed et al. \cite{elhoseny2019optimal} employed the Dragonfly and improved Firefly Algorithms (FFA) to categorize the images as normal and abnormal. This adjustment improved significantly due to this adjustment, as the peak signal to noise ratio (PSNR) reduced significantly. Melanoma diagnosis was enhanced utilizing the Whale Optimization Algorithm (WOA) and levy battles, as introduced in \cite{zhang2020skin}. Two datasets were analyzed using the developed structure, and the accuracy was 87 \% on both datasets.
Some of them suffer from premature convergence and local minima, especially when faced with a large solution space \cite{el2021clustering}. Often, this limit results in inefficient task scheduling solutions, which hurts system performance. Therefore, a globally optimal solution to the IoMT task scheduling problem is urgently needed. 

However, these existing approaches were still unable to achieve a high degree of efficiency. 
To overcome this problem, this paper aims to find the best solutions that lead for improving performance. Hence, we combine
transfer learning with meta-heuristic FS optimization to create an  available IoMT system. The characteristics of this system allow for outstanding performance, reasonable computing expenses, and address the financial concerns discussed earlier. As a result of the IoMT, it is necessary to treat and detect infection inside and outside the clinic. In order to use the system, internet-connected devices and a digital copy of scan. 
However, these existing approaches were still unable to achieve a high degree of efficiency. To overcome this problem, this paper aims to find the best solutions that lead for improving performance. \color{black} The main difference between the proposed model and previous approaches is that we combine transfer learning with meta-heuristic FS optimization to create an available IoMT system. The characteristics of this system allow for outstanding performance and reasonable computing expenses. Hence, this system is necessary to treat and detect infections and diseases  from anywhere. \color{black}

\section{Methodology} \label{pm}

\begin{figure}
    \centering
    \includegraphics[width=13cm]{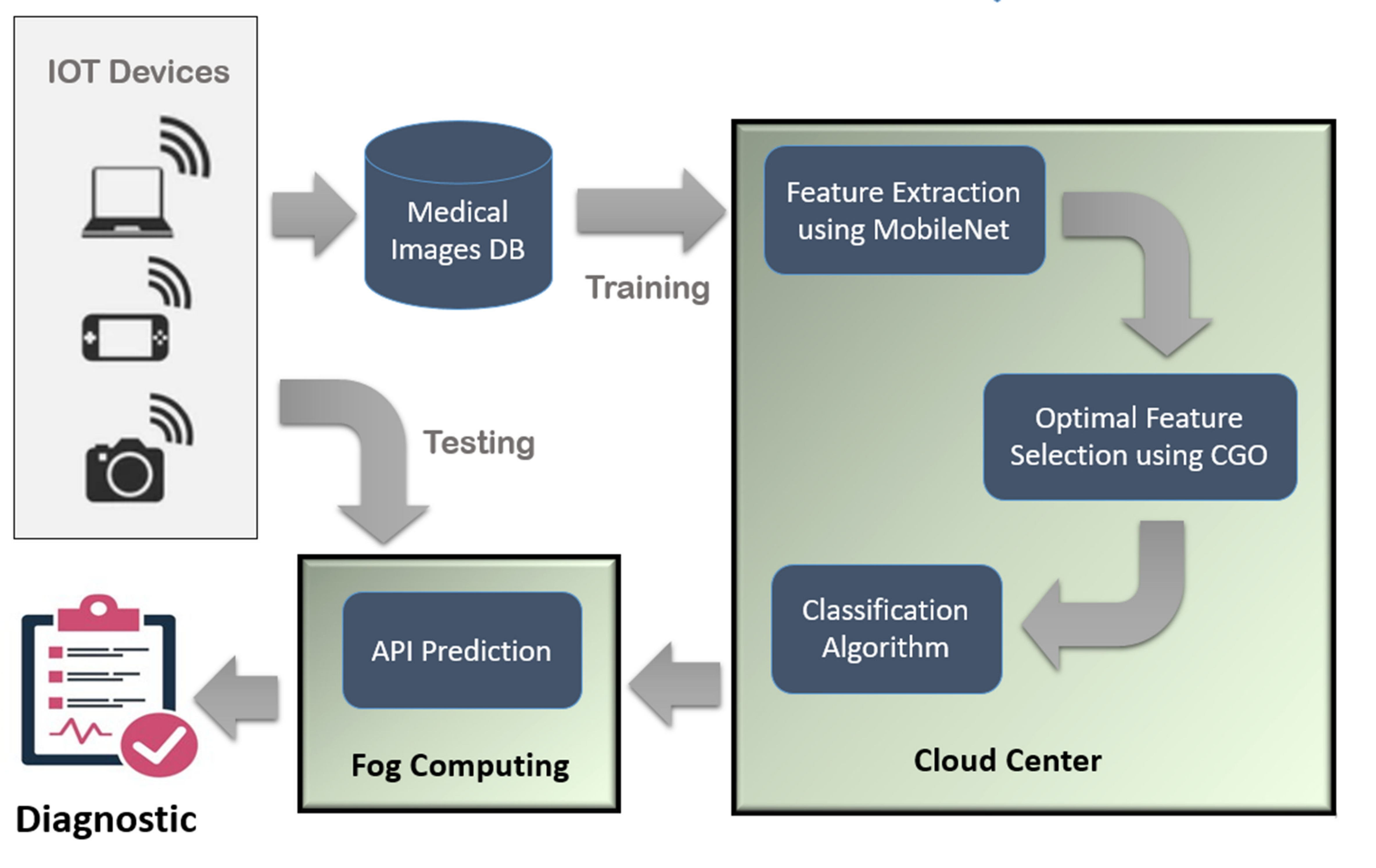}
    \caption{\label{fig:IoTframwork} Diagram of the proposed IoMT system. }
    
\end{figure}

Inside this field of medical image classification, detecting a user's illness using a medical database is an interesting topic. The present study used three datasets for image recognition analysis, with the major goal of achieving maximal performance in disease diagnosis. The three datasets investigated were ISIC-2016 \cite{gutman2016skin}, PH2 \cite{mendoncca2013ph} (Both for melanoma detection) and Blood-Cell classification \cite{liang2018combining}. Figure \ref{fig:IoTframwork} depicts the established IoMT's architecture. Initially, the IoMT devices capture medical images, and if the goal is to train the IoMT system, the image data could be sent to a cloud center. There still are three main processes at this level. Throughout the first stage, the features are extracted while using the TL architecture, as detailed in subsection \ref{SecFE}. The next stage is to find the relevant features using CGO. Lastly, the classification is performed, and the results can be dispersed across fog operating systems to save on communication costs if desired. If the goal is to identify the condition of the collected data, the training data in Fog operating systems are utilized. 


\subsection{Proposed IoMT system}

Our IoMT system is based on a computational cloud that communicates with a fog. Users may easily manipulate the data and parameters required to get the online service's classification results. This system component also handles communication between IoT devices (mobile phones and laptops) and the cloud center. Because the patient's images are all the same, the system can be used for various exams, proving its reliability. Image sizes, formats, and color conversions are adjusted as standards.

The IoMT system represented in Figure \ref{fig:IoTframwork} is what we offer to implement our methodology in the system in order to give a quick reaction and support the physician in making appropriate choices. There are two components in our system, cloud computing, and fog computing. 

These are done first by sending a medical image database to a training level in the cloud using IoT technologies. Using the training model, the created system from subsections \ref{SecFE} and \ref{SecFS} may be well. The pre-trained feature extraction technique is deployed on cloud service and benefits from the light and quick approach. There is well-known interoperability and limited resource use on embedded systems with the MobileNetV3 structure to extract the features. The introduced CGO algorithm, a lightweight and robust feature selection method, has been used upon feature extraction to minimize the features embedding set and only maintain the more essential features in each filtered image. We can speed up the training process by decreasing the number of features, which will allow us to arrive at a classification choice in an acceptable amount of time.

One of the two components included in this IoMT system is fog computing. It allows the approved training model to make predictions without re-training the system, saving time and reducing network traffic. As a result, fog computing devices can assist the expert in making a judgment on medical image diagnosis better than waiting for a choice from the cloud centers. In addition, the training process on the cloud centers is fine-tuned regularly, employing photos gathered from Connected devices and saved in a database. Thus, the training system's quality will improve, making better, more accurate decisions.

There will also be a web-based application that the transmitter can use to create a rapid forecast that uses the pre-trained or fine-tuning program to refine the system on a batch of new photographs. The sender will receive the final choice among other measurement metrics like accuracy to back up system forecasts.

\color{black}
\subsection{Feature Extraction Using TL}

This section gives a detailed description of the used transfer learning technique for features learning and extraction. As mentioned in section \ref{rw}, the pre-trained model for image classification tasks in computer vision is beneficial in training and inference speed. In addition, few parameters can be fine-tuned during the training process rather than training models from scratch. In our system, MobileNetV3 is used as the backbone of the feature extraction process where the top layers of the model are replaced with new layers, and only specific layers are fine-tuned. The MobileNetV3 is an optimized version generated by a network architecture search (NAS) algorithm called NetAdapt. The NetAdapt algorithms use MobileNetV1 and MobileNetV2 components to search for an optimal network architecture and kernel size to minimize the model size and latency alongside maximizing its performance.

\subsubsection{Efficient Deep Learning}

DL techniques and models have demonstrated success in various tasks, including image classification, image segmentation, and object detection \cite{tran2019video, ji2020action, liu2021ntire, ignatov2021real}. However, the challenges of these tasks, especially the quality and the impact of the learned representations, remain largely unexplored. Over the past decade, several DL architectures and training techniques have been proposed. For instance, researchers focus on exploiting the power of DL models to improve the model's performance and efficiency in terms of training time, computational resources, and accuracy. One of the most investigated DL models is convolutional neural networks with different architectures, designs, parameters, and training processes. Depthwise convolutions are DL components designed to exploit the spatial information in the input image and replace the traditional convolution layers, thus facilitating their deployment on embedded devices or edge applications. Various DL models have embraced the concept of depthwise convolutions to overcome the limitations of traditional convolution layers including MobileNets \cite{howard2017mobilenets, howard2019searching}, ShuffleNets \cite{zhang2018shufflenet}, NASNet \cite{zoph2018learning}, MnasNet \cite{tan2019mnasnet}, and EfficientNet \cite{tan2019efficientnet}. Unlike the traditional convolution layers, the depthwise convolution layers are used separately on each input channel. Thus, the models can be computationally inexpensive and trained with fewer parameters and less training time. In this section, we will focus on introducing the MobileNetV3 \cite{howard2019searching} and its core components. More detailed information will be discussed in the following sections, where we describe the MobileNetV3 as our feature extractor used in the proposed system.

Howard et al. \cite{howard2019searching} introduced the MobileNetV3 in two versions : MobileNetV3-large and MobileNetV3-small. The MobileNetV3 is designed to optimize the latency and accuracy of the previous version, which is the MobileNetV2 architecture. For instance, MobileNetV3-large improved the accuracy by 3.2\% compared to the MobileNetV2 while reducing the latency by 20\%. The MobileNetV3 was designed using a network architecture search (NAS) technique termed NetAdapt algorithm to search for the optimal network structure and kernel size of the depthwise convolution. As illustrated in Figure \ref{mb3_arch_fe} (Section \ref{SecFE}), the MobileNetV3 architecture is composed of the following core building blocks:
\begin{itemize}
    \item The depthwise separable convolutional layer has a depthwise convolutional kernel of size $3 \times 3$ followed by batch normalization and activation function.
    \item The $1 \times 1$ convolution (pointwise convolution) for linear combination computations of the depthwise separable convolutional layer and feature maps extraction.
    \item The global average pooling layer reduces the dimensionality of the feature maps.
    \item The inverted residual block inspired from the bottleneck blocks networks \cite{he2016deep} that use the residual skip connections mechanism. The inverted residual block consists of the following sub-blocks:
        \begin{itemize}
            \item The $1 \times 1$ expansion and projection convolutional layers with a depthwise convolutional kernel of size $1 \times 1$ to learn more complex representations and reduce the model's calculations.
            \item A depthwise separable convolutional layer.
            \item A residual skip connection mechanism. 
        \end{itemize}
    \item The Squeeze-And-Excite block (SE block) \cite{tan2019mnasnet} to select the relevant features on a channel-wise basis.
    \item The h-swish activation function \cite{ramachandran2017searching,elfwing2018sigmoid} which is used interchangeably with the ReLU (Rectified linear unit) activation function.
\end{itemize}

\color{black}

\subsubsection{Feature Extraction Module} \label{SecFE}

\begin{figure}
    \centering
    \includegraphics[width=0.95\textwidth]{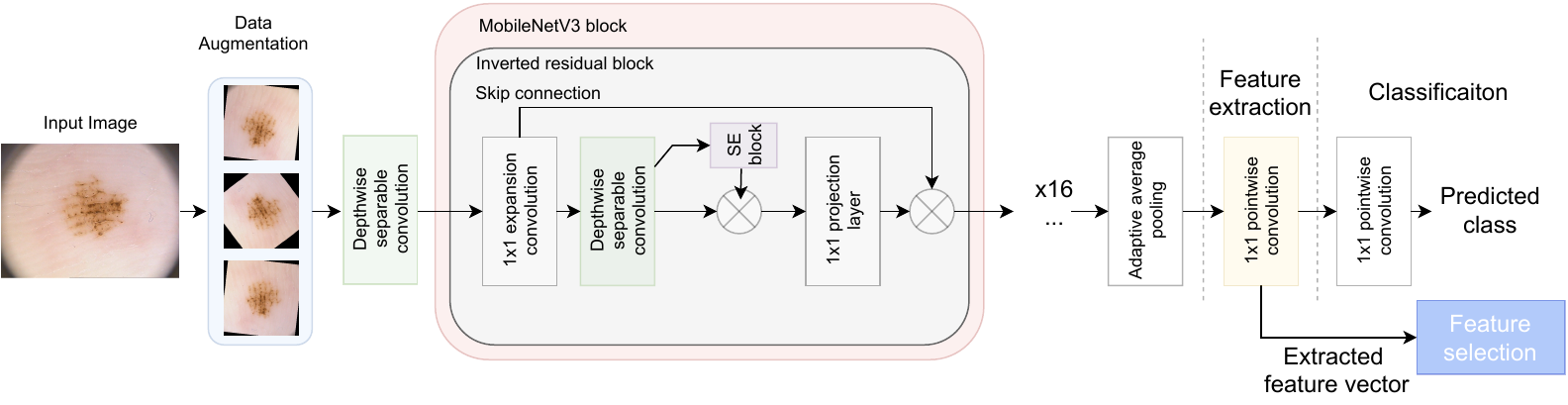}
    \caption{The building blocks of the proposed network architecture for feature extraction.}
    \label{mb3_arch_fe}
\end{figure}

Using different image datasets, the MobileNetV3 was fine-tuned to learn and extract feature vectors from inputted images of size $224 \times 224$. The MobileNetV3 was trained on the ImageNet dataset \cite{he2016deep}. In our experiments, the MobileNetV3-Large pre-trained model was employed and fine-tuned on the datasets having skin cancer and blood cells images. A $1 \times 1$ point-wise convolution (Conv) was used to replace the top layers used for classification in the MobileNetV3 model as shown in Figure \ref{mb3_arch_fe}. 

The $1 \times 1$ point-wise convolution can be seen as a multilayer perceptron (MLP) used for classification and feature extraction tasks. Thus, in our implementation, we used two $1 \times 1$ point-wise convolutions at the top of the model to extract features from the input images and fine-tune the model on the image classification task. Meanwhile, the MobileNetV3 building block consists of an inverted residual block inspired by the bottleneck blocks. The inverted residual block contains two important blocks: the depthwise separable convolution block and a Squeeze-And-Excite block used to link the input and output features on the same channels, thus improving the features representations with low memory usage. The depthwise separable convolution block consists of $3 \times 3$ depthwise convolution, batch normalization (BN), activation function, and $1 \times 1$ point-wise convolution where the order of execution of the layers is as follows: $(3 \times 3 Conv)\rightarrow(BN)\rightarrow(ReLU/h-swish)\rightarrow(1 \times 1 Conv)\rightarrow(BN)\rightarrow(ReLU/h-swish)$. In contrast, the Squeeze-And-Excite block consists of fully connected layers (FC) with non-linear transformation for global feature extraction using global pooling operation with the following execution order: $(Pool)\rightarrow(BN)\rightarrow(FC1)\rightarrow(ReLU)\rightarrow(FC2)\rightarrow(Sigmoid)$. Each building block can integrate a depthwise separable convolutional layer with different non-linearity function such as ReLU or hard swish (h-swish) which are defined in Equations \ref{eq:relu} and \ref{eq:hswich}, respectively.   

\begin{equation}
    \label{eq:relu}
    ReLU(x) = \max(0, x)
\end{equation}

\begin{equation}
    \begin{aligned}
        h\_swich(x) = x \cdot \sigma(x) \\
        \sigma(x) = \frac{ReLU6(x + 3)}{6}
    \end{aligned}
    \label{eq:hswich}
\end{equation}

Where $h\_swich$ is a modified version of the sigmoid activation function and $\sigma(x)$ defines the piece-wise linear complex analog function.

To extract the feature vector from each input image, we used the generated fine-tuned model on each dataset. We flattened the $1 \times 1$ point-wise convolutional layer (placed before the classification layer) output and used the output as the feature vector. The extracted feature vector for each image of size 128 will be fed into the feature selection process in the proposed system. 
The model was fine-tuned for 100 epochs with a batch of size 32 on each data set to produce the best classification performance. Meanwhile, to update the model's weight and bias parameters, we used the RMSprop optimizer with a learning
rate of $1e-4$. \color{black} To overcome the model's over-fitting, we used the dropout layer with a probability of 0.38.\color{black}

\color{black}
\subsection{Feature Selection Optimization}

During using methods for extraction of features, including MobileNetV3, the extracted features were not transmitted straight to the classification algorithm since it needed more processing time to reach. Feature Selection (FS) techniques reduce redundant or unusable features from retrieved patient data like a content decomposition method. It means that the FS process minimizes the quantity of data transferred. As a result, an optimized feature choice process was implemented wherein most of the critical features were defined using the optimizer, i.e., Chaos Game Optimization (CGO). 

\subsubsection{Chaos Game Optimization (CGO)}
\noindent

As a result of certain principles of chaos theory, the CGO relies on fractal self-similarity issues \cite{talatahari2021chaos}. According to the chaos theory, small changes in the early conditions of a chaotic system can significantly impact its future because of the system's dependence on its beginning conditions. Following this theory, the present state of a system can predict its future state, while the estimated existing state of the system does not identify its future state. In mathematics, the chaos game is constructing fractals by utilizing the main polygon pattern and a chosen randomly crucial point to create fractal patterns. The main goal is to construct a combination of points with a recurrent attitude to achieve a shape with a similar style in different ranges.

\begin{figure}
    \centering
    \includegraphics[width=13cm]{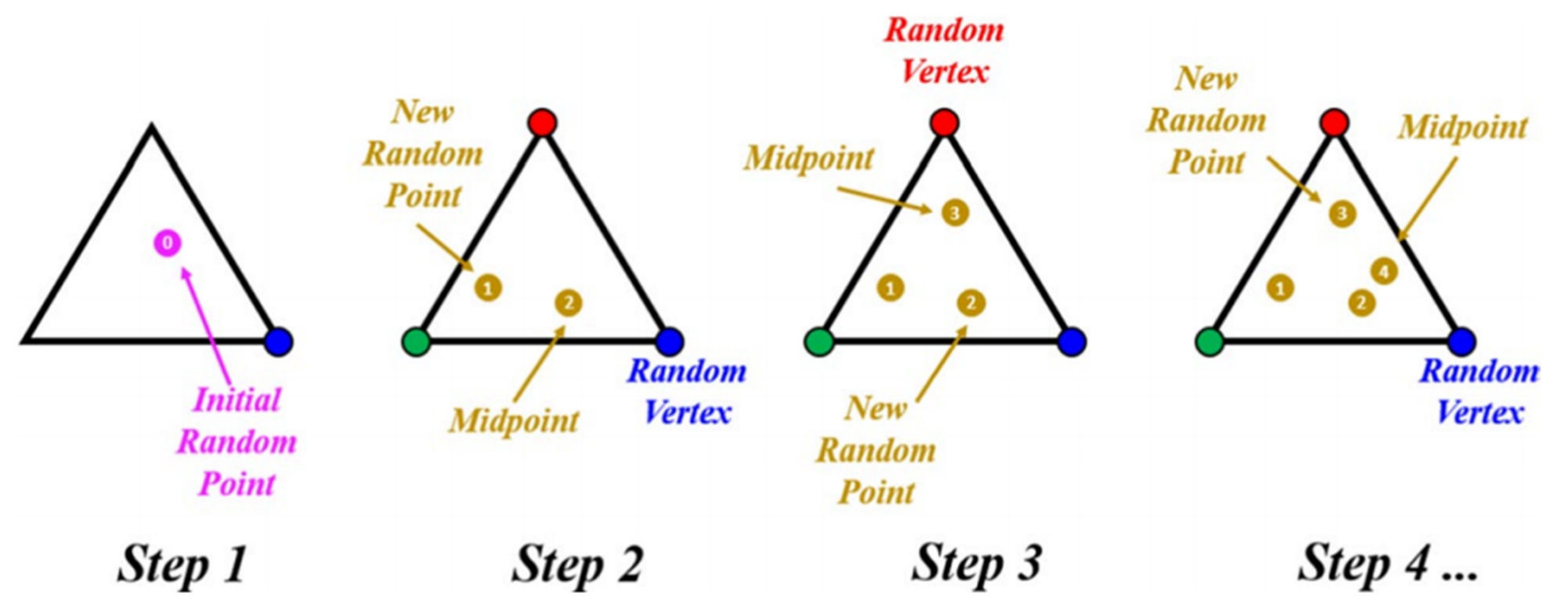}
    \caption{\label{fig:Triangle} Using chaotic game to create Sierpinski triangle \cite{talatahari2021chaos}. }
    
\end{figure}

Using a Sierpinski triangle fractal as an example, we may better appreciate the chaos game's theory. As shown in Figure \ref{fig:Triangle}, if three points are chosen for the main fractal structure, in this case, the output is the triangle. Selected vertices have been highlighted in red, green, or blue. The die utilized should have two red sides, two blue sides, and two green sides in this situation. First, a random point is chosen as the fractal's seed. A seed is moved from its starting location to the vertex corresponding to that color on each die roll by rolling it again and using its new location as a starting point for further reiterations. Finally, a dice is rolled multiple times before the Sierpinski triangle appears.

As a result of using the chaos game mechanics and fractals, the CGO method has been developed. Many candidate solutions ($S$) represent a few of Sierpinski's valid points. There are some choice factors ($s^j_k$) associated with each solution candidate ($S_{k}$). These selection factors reflect the placement of such eligible seeds within the Sierpinski triangle. The triangle can also be used to seek solutions.

The primary strategy is to generate new seeds in the search area that could be the newly eligible seeds by generating temporary triangles. Towards achieving this goal, four different approaches are described. There is an iteration of this technique across all eligible seeds and the $k^{th}$ temporary triangles inside the search domain. The triangle has three nodes inside the search area, including three $k^{th}$ initial points, the blue ($S_{k}$), green ($G$), and red points ($M_k$). In this temporary triangle, a die is used to create new seeds using the chaotic method. Chaos game principles are used in this temporary triangle, creating new points with a die and three seeds. The three seeds ($S_{k}$, the $G$ and $M_k$) are placed in order of importance, from first to third, respectively. When it comes to $S_k$'s first seed, a die with six faces (i.e., three red and three green) is used. Depending on the color of the die, the point is transformed in $S_{k}$ towards $M_k$ (red side) or $G$ (green side). When rolling dice comes up green/red, the point is moved over to either $G$/$M_k$. It is possible to replicate this feature by using a random number generation function that creates only two values [0 and 1] for the possibility of selecting red or green sides. The green side indicates that the seed placed in $S_{k}$ has moved to the $G$, while the red side indicates that the seed placed in $S_{k}$ has moved to $M_k$. Unaffected by the fact that both sides of the game are equally likely to emerge, creating two random numbers for both $M_k$ and $G$ assumes that perhaps the seed contained in $M_k$ is relocated anywhere along connected connections between the $M_k$ and $G$. As a result of the chaos game technique that manages this feature, some randomly generated factorials are also used based on the actuality of the seeds' movement inside the search region. The first point has the following mathematical:

\begin{equation}\label{Eq_S1}
P^1_i=S_k + \alpha_k\times(\beta_k\times G - \gamma_k\times M_k), \,    k=1,2,...,D.
\end{equation}

Where $S_k$ is the solution candidate ($kth$), $G$ refers to the global solution implemented so far. As the name suggests, $M_k$ is the average number of beginning points considered three points in the $kth$ temporary triangle. Seed motion limitations are modeled using the randomly generated factorial, where $\alpha_k$ is the seed's motion limitations. If wanting to represent the likelihood of rolling a dice, $\beta_k$ and $\gamma_k$ correspond to random integers of 0 or 1. $D$ is the number of eligible points (solution candidates).

Regarding the second point, which is placed in the $G$, a die with six faces (i.e., three red and three blue) is utilized. The point in the $G$ is moved to the $S_k$ (blue face) or the $M_k$ (red face). When a random number production function generates only two numbers, 0 and 1, for the possibility of picking red/blue faces, this property can be represented. When the blue face shows, the position of the seed in the $G$ is changed to the $S_k$. When the red face shows, the position of the point in the $G$ is changed to the $M_k$. Although each blue or red side has an equal chance to happen, the potential of generating two random numbers of 1 for $S_k$ and $M_k$ is also assumed that the point placed in $G$ is relocated along the course of the connected connections between $M_k$ and $S_k$. According to the chaotic game technique, transportation inside the search region should be limited based on the actuality of the seed; certain randomly generated factorials are used to manage this feature. The mathematical presentation for the second seed is as follows:

\begin{equation}\label{Eq_S2}
P^2_k=G + \alpha_k\times(\beta_k\times S_k - \gamma_k\times M_k), \,    k=1,2,...,D.
\end{equation}

In addition, for the third seed, which is placed in $M_k$, a die with three blue sides and three green sides is used. The seed in $M_k$ is transferred to the $S_k$ (blue side) or the $G$ (green side) by rolling the dice and relying on the color that shows green/blue. This functionality can be represented by a random integer creation function that generates only two values, 0 and 1, for the option of selecting green/blue faces. When the blue face shows, the position of the point in the $M_k$ is changed to the $S_k$. When the green face occurs, the place of the point in the $M_k$ is transferred to the $G$. Every one of the green and blue sides has an equal chance of occurring in this game. Then, creating two random numbers of $S_k$ and $G$. Next, the $M_k$ is transferred the path of the associated lines between the $G$ and the $S_k$. Based on the actuality of the point, movements inside this search region should be controlled using the chaotic game technique to regulate this feature; specific randomly generated factorials are used. The third seed has the following formula:

\begin{equation}\label{Eq_S3}
P^3_k=M_k + \alpha_k\times(\beta_k\times S_k - \gamma_k\times G), \,    k=1,2,...,D.
\end{equation}

The additional point is also used as a fourth point placed in the $S_k$ to conduct out all the stages of modification inside the search range. The technique for upgrading the fourth seed's placement is dictated by specific random fluctuations in the randomly chosen decision factors. The fourth seed has the following mathematical representation:

\begin{equation}\label{Eq_S4}
P^4_k=S_k(S^i_k=S^i_k+rand), \,    i=[1,2,...,N].
\end{equation}

Where, the points dimension is denoted by $N$. $i$ denotes an integer in the range [$1, N$]. $rand$ stands for an uniform random value [0, 1]. 

For managing and changing the rates of exploration and exploitation within the proposed CGO algorithm, four formulas are conducted to identify the $\alpha_k$ as shown in Eq. \ref{Eq_P2}, which is used to simulate the seeds mobility limitations. These four formulas are randomly employed to locate the position of the first through third seeds.

\begin{equation}\label{Eq_P2}
\alpha_k=\left\{ {\begin{array}{*{20}{l}}
R\\
2\times R\\
(\epsilon\times R)+1\\
(\epsilon\times R)+(\epsilon)\\
\end{array}} \right.
\end{equation}

where R denotes an uniform random value in the range [0,1]. Besides, $\epsilon$ and $\alpha$ are integers having random values ranged [0,1].

According to the self-similarity of the fractals, the early eligible seeds and the freshly formed seeds applying the chaos game principle must be considered to determine if the newly created seeds should be included or not with the total eligible seeds inside the search domain. As a result, the initial seeds are transformed into new individual points if they achieve the highest levels of self-similarity, or they are reserved if the new seeds achieve the lowest levels of self-similarity. Consider that the substitution operation is carried out in the mathematical technique to obtain a model with a reduced difficulty level. Since the Sierpinski triangle is a total form, the total points that have been found so far are used to complete its shape. If the solution variables ($S^j_k$) are out of bounds, it is crucial to deal with them as soon as they are discovered. $S^j_k$ is outside the range of variables in this example, and the flag advises adjusting the boundaries of those variables. After a predefined set of optimization rounds, the optimization method concludes.

\begin{algorithm}[H]
\caption{Algorithm of CGO}\label{Alg_CGO}
\begin{algorithmic}[1]
{
\State\textbf{Input:} \\
$D$: the number of starting eligible seeds. 
\State Initialize the starting positions ($S^j_k$) with random values of eligible seeds ($S_k$). 

\State\textbf{Output:} \\
$G$: the global best eligible seed. \\

\State \textbf{Method:}

\State Compute objective function for each eligible seed.

\Repeat
\For{$k=1$ to $D$}
    \State Create a Mean Group ($M_k$).
    \State Construct a temporary triangles on three vertices of $S_k$, $G$, and $M_k$
    \State Create new seeds by Eqs. \ref{Eq_S1} to \ref{Eq_S4}.
    
     \If {boundaries are crossed by new seeds}
    \State Position limitations can be adjusted for new seeds. 
    \EndIf
     \State Assess the fitness of new points. 
     \If {new seeds have higher objective function than the last initial eligible seeds}
    \State  Substitute the last points by the new ones. 
    \EndIf
    \If{the best solution is achieved}
    \State Amend $G$.
    \EndIf
   \EndFor
 \Until{The iteration criterion has been met.}
 }
 \State Return $G$.
\end{algorithmic}
\end{algorithm}


Algorithm \ref{Alg_CGO} outlines the steps of the CGO algorithm. Besides, Figure \ref{fig:Flowchart} depicts the flowchart of this algorithm. Initially, the beginning locations of the solution candidates (X) inside the search region are determined by random selection. Secondly, determine the initial solution options' objective value based on the initial seeds' self-similarity. Then, it produces the Global Best ($G$) pertinent to the seed with a high eligibility level. Furthermore, generate a Mean Group ($M_k$) that used a random choice technique for each eligible point ($S_k$) inside the search area. Also, create a temporary triangle with the required three vertices of $S_k$, $M_k$, and $G$ for each eligible seed ($S_k$) inside the search region. Subsequently, find four seeds for each temporary triangle using Eqs. \ref{Eq_S1}–\ref{Eq_S4}. Afterward, the $s^j_k$ external variables scope should be checked for boundary conditions. Moreover, self-similarity is taken into account while calculating the objective function of these new seeds. Finally, It is time to replace the initial eligible points with new seeds if their objective functions show high self-similarity levels. 
\color{black}

\subsubsection{Optimal Feature Selection} \label{SecFS}

\begin{figure}
    \centering
    \includegraphics[width=13cm]{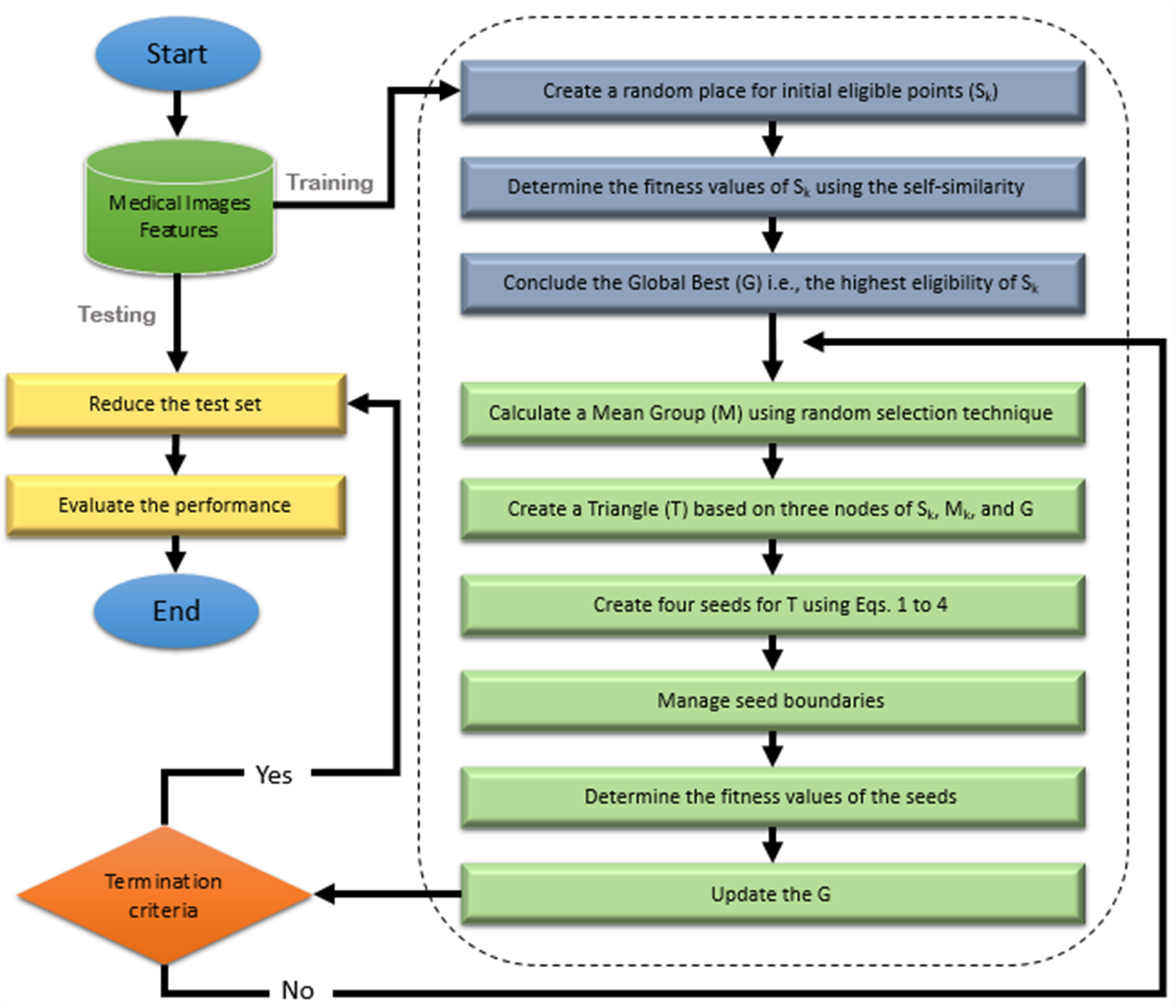}
    \caption{\label{fig:Flowchart} Flowchart showing the proposed methodology on the FS method. }
    
\end{figure}

In general, the feature extraction methods are separated into training and test datasets, with the training dataset used to learn the model to identify the essential features. 
Figure \ref{fig:Flowchart} depicts the stages of the binary CGO optimization technique. Firstly, the CGO is to produce a series of $N$ agents $X$ that depict the FS best solution. Then, the following formula is used to carry out a task:

\begin{equation}\label{eqp1}
    X_i=rand*(U-L)+L,\, i=1,2,...,N, \, j=1,2,...,Dim
\end{equation}

The dimension of the specific issue is denoted by $Dim$ in Eq. (\ref{eqp1}) (i.e., the number of features). In comparison, the search space is defined by $U$ and $L$.
A further step is to acquire the Boolean edition from each $X i$, which is accomplished to use the following equations:
\begin{equation}\label{Eq_P3}
BX_{ij}=\left\{ {\begin{array}{*{20}{c}}
1& if \ X_{ij}>0.5\\
0&otherwise\\
\end{array}} \right.
\end{equation}

The objective value from each $X i$ is computed by applying optimization technique, which depends on the binary $BX i$ and classifying mistakes.  
\begin{equation}\label{Eq_P4}
Fit_i=\lambda\times\gamma_i+(1-\lambda) \times \left(\frac{|BX_i|}{Dim}\right), 
\end{equation}

in which $(\frac{|BX_i|}{Dim})$ represents the ratio of defined feature sets. The classifying fault utilizing $SVM$ is denoted by $\gamma_i$. SVM is commonly used because it is much more steady than other classification techniques and has fewer parameters. In contrast, $\lambda$ is a measurement that always had to adjust the proportion of selected features and categorization fault.

The following step is to examine the halt criteria, and if they have been encountered, the best solution is brought back. Alternatively, the automatic update steps are repeated.


The classification is conducted after getting the optimal features from the CGO algorithm. We use a machine learning technique, such as stochastic gradient descent (SGD). To train deep neural networks with better prediction capabilities by investigating the top non-convex cost space is among the main objectives in DL. As a typical reason to describe this phenomenon, one can demonstrate that perhaps the cost landscape on its own is simple, with no misleading local optimal. However, it turns out that the cost landscape of superior DL models has fictitious local (or global) optimum, and Stochastic Gradient Descent (SGD) is capable of detecting them \cite{liu2019bad}.
Nevertheless, the SGD approach, launched at random, has high generalization qualities in the real world. In explaining this achievement, a hypothesis would have to provide for the entire method course, which became apparent. The problem remains challenging, even for the most advanced DL trained on datasets, which are still in the experimental stage.

\section{Experiments} \label{ex}

\subsection{Experimental data} \label{ed}

Three datasets of medical photos were used to conduct image classification task for our experimental tests: PH2 \cite{mendoncca2013ph}, ISIC-2016 \cite{gutman2016skin}, and Blood-Cell dataset as in \cite{liang2018combining}.

\begin{figure}
    \centering
    \includegraphics[width=13cm]{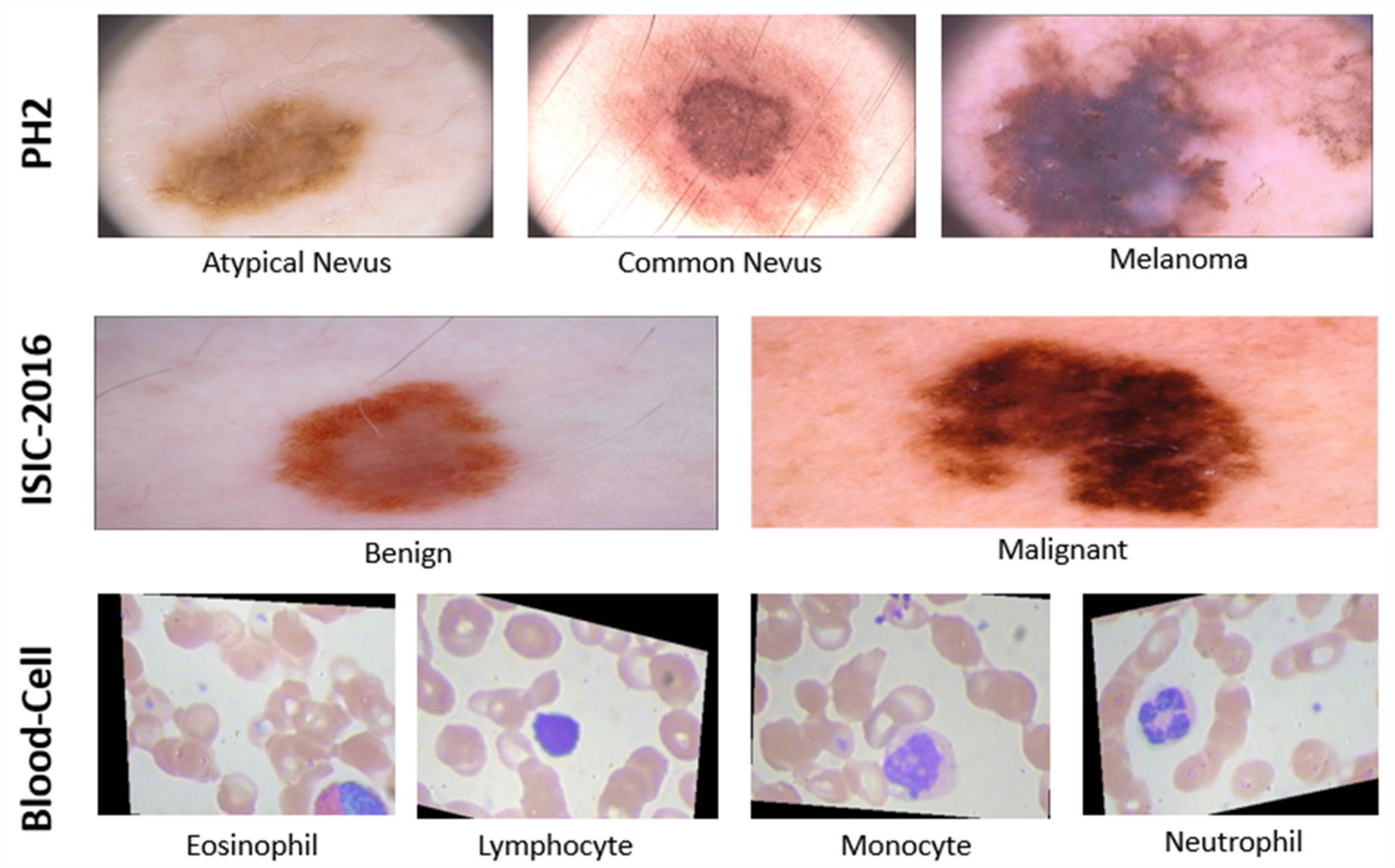}
    \caption{\label{fig:Examples}Example medical image samples for classification task from the three selected datasets. }
    
\end{figure}

(1) \textbf{PH2}: A total of 200 dermoscopic images were included in this dataset, including 80 Atypical Nevus, 80 Common Nevus, and 40 Melanoma. This dataset can be freely downloaded at \url{http://www.fc.up.pt/addi/ph2\%20database.html}. Table \ref{tab:desc}  describes more detail about each dataset and its respective classes. As an example, Figure \ref{fig:Examples} shows some of the image samples from the selected databases.

(2) \textbf{ISIC-2016}: In total, 1179 photos are included in this dataset, which is separated into two categories: Most of the data is benign, whereas the remainder is malignant. There is a link on the website to get this database \url{https://challenge.isic-archive.com/data}. 

(3) \textbf{Blood-Cell}: The dataset is collected from publicly available dataset from BCCD Dataset (\url{https://www.kaggle.com/paultimothymooney/blood-cells/data}). It comprises 12,442 blood cell images, 2487 test sets, and 9957 training sets. These images are classified into four types of blood cells: Eosinophils, Lymphocytes, Monocytes, and Neutrophils. There are 2496 Eosinophils, 2484 Lymphocytes, 2477 Monocytes, and 2498 Neutrophils in the training set, while 623 Eosinophils, 623 Lymphocytes, 620 Monocytes, and 624 Neutrophils are in the testing set.

\begin{table}[t]
\centering
\caption{\label{tab:desc}Dataset Description.}
\scalebox{0.9}{
\begin{tabular}{|c| c| c |c |c|}
\hline
\textbf{Dataset name}	& \textbf{Class}	& \textbf{Training data}	& \textbf{ Test data}	&\textbf{ \# Images per class}\\\hline\hline

\textit{$Ph^2$}	& Common Nevus	& 68	& 12	& 80\\
&Atypical Nevus&	68 &	12	& 80\\
& Melanoma	& 34	& 6	& 40\\
&\textbf{Total }&	170&	30&	200 \\\hline

\textit{ISIC-2016}	& Malignant  	& 173	& 75	& 248 \\
& Benign 	& 727	& 304	& 1,031 \\
&\textbf{Total}&	900&	379&	1279 \\\hline

\textit{Blood-Cell}	
& Neutrophil	& 2499	& 624	& 3123\\
& Monocyte	& 2478	& 620	& 3098\\
&Lymphocyte&	2483 &	620	& 3103\\
& Eosinophil	& 2497	& 623	& 3120\\
&\textbf{Total }&	9957&	2487&	12444 \\\hline
\end{tabular}
}
\end{table}

\subsection{Evaluation Metrics} \label{em}

This research was evaluated Using the metrics in Table \ref{tab:metric}: balanced accuracy, accuracy, recall, precision, and F1-score. Balanced accuracy is specified as the average accuracy acquired across all classes. The quantity created across all predicted values is referred to as accuracy. The recall is calculated as the proportion of actual numbers to values that should have been predicted. Precision is calculated as the proportion of actual numbers to defined properties. Finally, the F1-score indicates a class imbalance between Recall and Precision. 

Where False Positives (FP) refers to the precise number of positives discovered from actual samples, when referring to True Negatives (TN), it refers to the correct number of non-modular data found. Besides, the number of nodular data discovered in a non-nodular sample is known as False Positives (FP). Finally, it represents the number of faults identified in actual nodular data, referred to as False Negatives (FN).

\begin{table}[]
\centering
\caption{\label{tab:metric} Various performance parameters.}

\begin{tabular}{|c |c|}
\hline
\textbf{Metrics} & \textbf{Formula} \\ \hline 
    Recall & \parbox{7cm}{\begin{equation*}  \frac{TP}{TP+FN} \label{r}\end{equation*} }        \\ \hline
    Precision & \parbox{7cm}{\begin{equation*} \frac{TP}{TP+FP} \label{p} \end{equation*}}          \\ \hline
    Accuracy  & \parbox{7cm}{\begin{equation*} \frac{TP+TN}{TP+TN+FP+FN} \label{ac} \end{equation*}}              \\ \hline
    F1-Score & \parbox{7cm}{\begin{equation*} \frac{2*Precision*Recall}{Precision+Recall}  \label{f1}  \end{equation*}}              \\ \hline
    Sensitivity & \parbox{7cm}{\begin{equation*} \frac{TP}{TP+FN}    \label{Sensitivity}  \end{equation*}}              \\ \hline
    Specificity & \parbox{7cm}{\begin{equation*} \frac{TN}{FP+TN} \label{Specificity}  \end{equation*}}              \\ \hline
    Balanced Accuracy & \parbox{7cm}{\begin{equation*} \frac{Sensitivity + Specificity}{2} \label{bc}  \end{equation*}}              \\ \hline

\end{tabular}
\end{table}


\subsection{Experimental results and analysis} \label{er}

The results analysis and discussion of experiments for the suggested approach task scheduling technique are presented in this section. First, we compare our approach with various meta-heuristic optimization strategies. Afterward, the three classifiers are compared, namely, k-Nearest Neighbor (kNN), Support Vector Machines (SVM), and stochastic gradient descent (SGD). Then, we compare our results to those of other current medical image classification algorithms. Finally, A comparison with published techniques has been conducted.

To objectively examine the effectiveness of our proposed approach, we compared it to nine well-known algorithms. The metaheuristic optimizers, in particular, 

\begin{itemize} 

\item Particle Swarm Optimization (PSO) \cite{eberhart1995new}, 
\item Multi-Verse Optimizer (MVO) \cite{mirjalili2016multi}, 
\item Grey Wolf Optimization (GWO) \cite{emary2016binary}, 
\item Moth-Flame Optimization (MFO)\cite{mirjalili2015moth}, 
\item Whale Optimization Algorithm (WOA) \cite{mirjalili2016whale}, 
\item Firefly Algorithm (FFA) \cite{yang2010firefly}, 
\item Bat Algorithm (BAT) \cite{yang2010new}, and 
\item Hunger Games Search (HGS) \cite{yang2021hunger}.

\end{itemize}

\begin{table}[]
\centering
  \caption{The parameters of each FS optimizer and their values.}
\begin{tabular}{|c|c|l|c|}
\hline
\textbf{S\#}            & \textbf{Optimizer}       & \multicolumn{1}{c|}{\textbf{Parameter}} & \textbf{Value} \\ \hline
\multirow{3}{*}{1}      & \multirow{3}{*}{PSO}     & Vmax                                    & 6.0            \\ \cline{3-4} 
                        &                          & Wmax                                    & 0.9            \\ \cline{3-4} 
                        &                          & Wmin                                    & 0.2            \\ \hline
\multirow{2}{*}{2}      & \multirow{2}{*}{MVO}     & WEPMin                                  & 0.2            \\ \cline{3-4} 
                        &                          & WEPMax                                  & 1.0            \\ \hline
\multirow{2}{*}{3}      & \multirow{2}{*}{GWO}     & a                                       & 2.0            \\ \cline{3-4} 
                        &                          & r                                       & {[}-1,1{]}     \\ \hline
\multirow{2}{*}{4}      & \multirow{2}{*}{MFO}     & B                                       & 1.0            \\ \cline{3-4} 
                        &                          & L                                       & {[}-1,1{]}     \\ \hline
\multirow{2}{*}{5}      & \multirow{2}{*}{WOA}     & a                                       & 2.0            \\ \cline{3-4} 
                        &                          & r                                       & 1.0            \\ \hline
\multirow{3}{*}{6}      & \multirow{3}{*}{FFA}     & Alpha                                   & 0.5            \\ \cline{3-4} 
                        &                          & BetaMin                                 & 0.2            \\ \cline{3-4} 
                        &                          & Gamma                                   & 1.0            \\ \hline
\multirow{2}{*}{7}      & \multirow{2}{*}{BAT}     & QMin                                    & 0.0            \\ \cline{3-4} 
                        &                          & QMax                                    & 2.0            \\ \hline
\multirow{4}{*}{8}      & \multirow{4}{*}{HGS}     & VC2 = 0.03                              & 0.0            \\ \cline{3-4} 
                        &                          & Vmax                                    & 6.0            \\ \cline{3-4} 
                        &                          & Wmax                                    & 0.9            \\ \cline{3-4} 
                        &                          & Wmin                                    & 0.2            \\ \hline
\multicolumn{1}{|c|}{9} & \multicolumn{1}{c|}{CGO} &       { $\beta$ and $\gamma$   }                              & [1, 2]               \\ \hline
\end{tabular}
\label{tab:Param}
\end{table}

As seen in Table \ref{tab:Param}, each optimizer retains a particular set of parameters. 
As the number of search agents increases, so does the likelihood of finding a worldwide optimal. The sample size is set at 50 in all experiments. The number of search agents could be reduced complexity.

The nine optimization techniques were combined with standard machine learning classifiers to produce the findings, such as KNN, SVM, and SGD. (a) According to KNN, an unidentified sample's classification is determined by the geographical sharing of benefits in that population. We can then find out where the k closest examples are located. The length among items is used to determine consistency. A typical length in a Euclidean distance is based on a mathematical formula. (b) It is possible to use SVMs as classification algorithms by altering the distributed space of data. SVM uses statistical knowledge for the classification task. Calculating hyper-plane can be used to understand statistics. The hyper-plane is defined based on the kernel used during a plot. Linear, polynomial, and RBF kernels are among the most common kernel types. (c) There are many advantages to using the SGD technique. An explanation for such success had to cover a broad duration of the procedure, which became apparent. Only the most robust DL learned on data, already in the test stage, have difficulty solving the challenge.

\subsubsection{Analysis Results}

When evaluating these optimization techniques, multiple measures are used. Evaluation of each method was based on Recall, Precision, Accuracy, and F1 Score. PH2, ISIC-2016, and Blood-Cell dataset are represented in \ref{tab:resultsPH2}, \ref{tab:resultsISIC}, and \ref{tab:resultsBlood}, respectively. In these tables, the bolded results are the highest accurate ones. According to the outcomes shown in these tables, the SGD-based CGO beats PSO, MVO, GWO, MFO, WOA, FFA, BAT, and HGS.

\begin{table}[]
\centering
\caption{Results of each algorithms on PH2 dataset.}

\begin{tabular}{|c|l|c|c|c|c|c|}
\hline
\multicolumn{1}{|l|}{Optimizer} &
  Classifier &
  \multicolumn{1}{l|}{Recall} &
  \multicolumn{1}{l|}{Precision} &
  \multicolumn{1}{l|}{F1 Score} &
  \multicolumn{1}{l|}{Accuracy} &
  \multicolumn{1}{l|}{Balanced Accuracy} \\ \hline
\multirow{3}{*}{PSO}    & SGD & 0.9714 & 0.9719 & 0.9714 & 0.9714 & 0.9762 \\ \cline{2-7} 
                        & KNN & 0.9564 & 0.9569 & 0.9565 & 0.9564 & 0.9732 \\ \cline{2-7} 
                        & SVM & 0.9679 & 0.9684 & 0.9679 & 0.9679 & 0.9702 \\ \hline
\multirow{3}{*}{MVO}    & SGD & 0.9750 & 0.9753 & 0.9750 & 0.9750 & 0.9792 \\ \cline{2-7} 
                        & KNN & 0.9561 & 0.9566 & 0.9562 & 0.9561 & 0.9762 \\ \cline{2-7} 
                        & SVM & 0.9679 & 0.9684 & 0.9679 & 0.9679 & 0.9702 \\ \hline
\multirow{3}{*}{GWO}    & SGD & 0.9749 & 0.9751 & 0.9750 & 0.9751 & 0.9792 \\ \cline{2-7} 
                        & KNN & 0.9719 & 0.9721 & 0.9716 & 0.9715 & 0.9762 \\ \cline{2-7} 
                        & SVM & 0.9678 & 0.9694 & 0.9679 & 0.9679 & 0.9702 \\ \hline
\multirow{3}{*}{MFO}    & SGD & 0.9750 & 0.9753 & 0.9750 & 0.9750 & 0.9792 \\ \cline{2-7} 
                        & KNN & 0.9714 & 0.9719 & 0.9714 & 0.9714 & 0.9762 \\ \cline{2-7} 
                        & SVM & 0.9679 & 0.9684 & 0.9679 & 0.9679 & 0.9702 \\ \hline
\multirow{3}{*}{WOA}    & SGD & 0.9714 & 0.9718 & 0.9715 & 0.9714 & 0.9732 \\ \cline{2-7} 
                        & KNN & 0.9571 & 0.9576 & 0.9572 & 0.9571 & 0.9762 \\ \cline{2-7} 
                        & SVM & 0.9714 & 0.9719 & 0.9714 & 0.9714 & 0.9762 \\ \hline
\multirow{3}{*}{FFA}    & SGD & 0.9679 & 0.9684 & 0.9679 & 0.9679 & 0.9702 \\ \cline{2-7} 
                        & KNN & 0.9564 & 0.9569 & 0.9565 & 0.9564 & 0.9762 \\ \cline{2-7} 
                        & SVM & 0.9714 & 0.9719 & 0.9714 & 0.9714 & 0.9762 \\ \hline
\multirow{3}{*}{BAT}    & SGD & 0.9750 & 0.9753 & 0.9750 & 0.9750 & 0.9792 \\ \cline{2-7} 
                        & KNN & 0.9561 & 0.9566 & 0.9562 & 0.9561 & 0.9762 \\ \cline{2-7} 
                        & SVM & 0.9714 & 0.9719 & 0.9714 & 0.9714 & 0.9762 \\ \hline
\multirow{3}{*}{HGS}    & SGD & 0.9750 & 0.9753 & 0.9750 & 0.9750 & 0.9792 \\ \cline{2-7} 
                        & KNN & 0.9564 & 0.9569 & 0.9565 & 0.9564 & 0.9732 \\ \cline{2-7} 
                        & SVM & 0.9714 & 0.9719 & 0.9714 & 0.9714 & 0.9762 \\ \hline
\multirow{3}{*}{CGO}    & SGD &  \textbf{0.9751} &  \textbf{0.9754} &  \textbf{0.9751} &  \textbf{0.9752} &  \textbf{0.9793} \\ \cline{2-7} 
                        & KNN & 0.9750 & 0.9753 & 0.9750 & 0.9750 & 0.9792 \\ \cline{2-7} 
                        & SVM & 0.9714 & 0.9719 & 0.9714 & 0.9714 & 0.9762 \\ \hline
\end{tabular}

\label{tab:resultsPH2}
\end{table}

On the PH2 dataset, Table \ref{tab:resultsPH2} shows that the CGO approach plays a significant role in feature selection when applying an SGD classifier since the findings are still effective; this is apparent throughout all measures. Analyzing results on the accuracy metric, using the SGD classifier, CGO can classify 97.52 \% of the test set, which is higher than the findings of the other optimization algorithms. According to the CGO, the BAT, HGS, MVO, MFO, and GWO in the second level are both at 97.50 \%. Moreover, the PSO's accuracy results are on par with WOA's, at 97.14 \%. Lately, the FFA's result has been the worst performance (i.e., 96.79 \%). On another view, The CGO achieved 97.54 \% on the precision metric, which was the best result on the SGD algorithm. BAT, HGS, MFO, and MVO came in the second level, 97.53 \%. They are followed by the GWO, which achieved 97.51 \%. Then, With the same level of precision. PSO and DLOHGS both have 97.19 \%. Last but not least, FFA has the lowest performance with 96.84 \%. To make things even better, the recall measure for the SGD classifier was 97.51 \% for CGO, 97.50 \% for HGS,  MVO, MFO, and BAT, 97.49 \% for GWO, 97.14 \% for PSO, and WOA, and 96.79 \% for FFA. In terms of F1-score, our CGO algorithm came out on top, with 97.51 \%. CGO is followed by the BAT, HGS, MFO, MVO, and GWO algorithms, 97.50 \%. Also, WOA achieved 97.15 \%. The last algorithms, PSO and FFA, are the worst in performance. In addition, the balanced accuracy measure for our CGO algorithm was 97.93 \%. Following CGO are BAT, MVO, MFO, GWO, and HGS algorithms with 97.92 \% each. More than that, the PSO has 97.62 \% accuracy. Lastly, FFA and WOA had the worst results with 97.32 \% and 97.02 \%, respectively. However, integrating these nine optimization techniques with the KNN classifier and SVM classifier produced the lowest metrics results compared with the SGD classifier.

\begin{table}[]
\centering
\caption{Results of each algorithms on ISIC-2016 dataset.}

\begin{tabular}{|c|l|c|c|c|c|c|}
\hline
\multicolumn{1}{|l|}{Optimizer} &
  Classifier &
  \multicolumn{1}{l|}{Recall} &
  \multicolumn{1}{l|}{Precision} &
  \multicolumn{1}{l|}{F1 Score} &
  \multicolumn{1}{l|}{Accuracy} &
  \multicolumn{1}{l|}{Balanced Accuracy} \\ \hline
\multirow{3}{*}{PSO}    & SGD & 0.8575 & 0.8482 & 0.8390 & 0.8575 & 0.6852 \\ \cline{2-7} 
                        & KNN & 0.8657 & 0.8569 & 0.8523 & 0.8657 & 0.7072 \\ \cline{2-7} 
                        & SVM & 0.8654 & 0.8570 & 0.8587 & 0.8654 & 0.7454 \\ \hline
\multirow{3}{*}{MVO}    & SGD & 0.8470 & 0.8389 & 0.8418 & 0.8470 & 0.7288 \\ \cline{2-7} 
                        & KNN & 0.8633 & 0.8539 & 0.8498 & 0.8633 & 0.7121 \\ \cline{2-7} 
                        & SVM & 0.8654 & 0.8570 & 0.8587 & 0.8654 & 0.7454 \\ \hline
\multirow{3}{*}{GWO}    & SGD & 0.8443 & 0.8392 & 0.8414 & 0.8443 & 0.7372 \\ \cline{2-7} 
                        & KNN & 0.8391 & 0.8310 & 0.8341 & 0.8391 & 0.7189 \\ \cline{2-7} 
                        & SVM & 0.8681 & 0.8598 & 0.8610 & 0.8681 & 0.7470 \\ \hline
\multirow{3}{*}{MFO}    & SGD & 0.7995 & 0.8015 & 0.8005 & 0.7995 & 0.6892 \\ \cline{2-7} 
                        & KNN & 0.8364 & 0.8317 & 0.8338 & 0.8364 & 0.7273 \\ \cline{2-7} 
                        & SVM & 0.8681 & 0.8598 & 0.8610 & 0.8681 & 0.7470 \\ \hline
\multirow{3}{*}{WOA}    & SGD & 0.8391 & 0.8399 & 0.8395 & 0.8391 & 0.7490 \\ \cline{2-7} 
                        & KNN & 0.8678 & 0.8605 & 0.8531 & 0.8678 & 0.7139 \\ \cline{2-7} 
                        & SVM & 0.8681 & 0.8598 & 0.8610 & 0.8681 & 0.7470 \\ \hline
\multirow{3}{*}{FFA}    & SGD & 0.8470 & 0.8378 & 0.8204 & 0.8470 & 0.6485 \\ \cline{2-7} 
                        & KNN & 0.8654 & 0.8570 & 0.8514 & 0.8654 & 0.7238 \\ \cline{2-7} 
                        & SVM & 0.8681 & 0.8598 & 0.8610 & 0.8681 & 0.7470 \\ \hline
\multirow{3}{*}{BAT}    & SGD & 0.8760 & 0.8775 & 0.8579 & 0.8760 & 0.7068 \\ \cline{2-7} 
                        & KNN & 0.8670 & 0.8601 & 0.8520 & 0.8670 & 0.7206 \\ \cline{2-7} 
                        & SVM & 0.8654 & 0.8570 & 0.8587 & 0.8654 & 0.7454 \\ \hline
\multirow{3}{*}{HGS}    & SGD & 0.8707 & 0.8622 & 0.8614 & 0.8707 & 0.7386 \\ \cline{2-7} 
                        & KNN & 0.8649 & 0.8565 & 0.8510 & 0.8649 & 0.7272 \\ \cline{2-7} 
                        & SVM & 0.8760 & 0.8684 & 0.8680 & 0.8760 & 0.7520 \\ \hline
\multirow{3}{*}{CGO}    & SGD &  \textbf{0.8839} &  \textbf{0.8781} &  \textbf{0.8751} &  \textbf{0.8839} &  \textbf{0.7569} \\ \cline{2-7} 
                        & KNN & 0.8311 & 0.8233 & 0.8265 & 0.8311 & 0.7089 \\ \cline{2-7} 
                        & SVM & 0.8628 & 0.8544 & 0.8564 & 0.8628 & 0.7437 \\ \hline
\end{tabular}

\label{tab:resultsISIC}
\end{table}

The proposed CGO algorithm outperformed other optimization techniques on the ISIC-2016 dataset, as seen in Table \ref{tab:resultsISIC}. The accuracy of the CGO algorithm for the SGD classifier was 88.39 \%, the best performance. In comparison, the BAT was at the second level, with 87.60 \%. With 87.07 \% of the vote, the HGS algorithm follows the preceding two. The PSO algorithm, which has 85.75 \%, follows the preceding three methods. The FFA and MVO algorithms (84.77 \%) are similar to their predecessors' algorithms. The algorithms that follow are the GWO (84.43 \%), WOA (83.91 \%), and MFO (79.95 \%). For the precision measure, our suggested CGO approach achieved a score of 87.81 \%. Following the FFA comes the HGS, which has an 87.75 \% rating. It was 86.22 \% for the HGS algorithm to keep up with them. 84.82 \% and 83.99 \% are the relative percentages for the PSO and WOA algorithms after the previous two algorithms in order of importance. The previous algorithms are followed by GWO, MVO, and FFA, which have respective success rates of 83.92 \%, 83.99 \%, and 83.78 \%. The MFO, on the other hand, has the lowest performance of 80.15 \%. As a result of the recall metric, 88.39 \% of the test samples were able to be compared using CGO, BAT, HGS, PSO, FFA, MVO, and GWO algorithms, while 83.91 \% of them were compared using the WOA method and 79.95 \% were compared using the MFO algorithm. For example, the proposed CGO outscored previous algorithms by 87.51 \% on the F1-Score scale. 86.14 \% was obtained by HGS, which HGS followed. Next, BAT, MVO, GWO, and WOA have 85.79 \%, 84.18 \%, 84.14 \%, and 83.95 \%, respectively. Finally, MFO gets the poorest performance with 80.05 \%, but not the latest. There was a 75.69 \% balanced accuracy of the CGO algorithm, the best performance. Regarding the WOA's and the HGS's performance in the second and third levels, respectively, they scored 74.90 \% and 73.86 \%. GWO is behind with 73.72 \%. FFA scored 64.85 \%, the lowest possible score.

\begin{table}[ht!]
\centering
\caption{Results of each algorithms on Blood-Cell dataset.}

\begin{tabular}{|c|c|c|c|c|c|c|}
\hline
\multicolumn{1}{|l|}{Optimizer} & Classifier & Recall          & Precision       & F1 Score        & Accuracy        & Balanced   Accuracy \\ \hline
\multirow{3}{*}{PSO}               & SGD        & 0.8862          & 0.9104          & 0.8888          & 0.8862          & 0.8862              \\ \cline{2-7} 
                                   & KNN        & 0.8866          & 0.9100          & 0.8890          & 0.8866          & 0.8866              \\ \cline{2-7} 
                                   & SVM        & 0.8862          & 0.9102          & 0.8888          & 0.8862          & 0.8862              \\ \hline
\multirow{3}{*}{MVO}               & SGD        & 0.8870          & 0.9106          & 0.8895          & 0.8870          & 0.8870              \\ \cline{2-7} 
                                   & KNN        & 0.8858          & 0.9094          & 0.8883          & 0.8858          & 0.8858              \\ \cline{2-7} 
                                   & SVM        & 0.8866          & 0.9109          & 0.8892          & 0.8866          & 0.8866              \\ \hline
\multirow{3}{*}{GWO}               & SGD        & 0.8874          & 0.9102          & 0.8898          & 0.8874          & 0.8874              \\ \cline{2-7} 
                                   & KNN        & 0.8854          & 0.9093          & 0.8880          & 0.8854          & 0.8854              \\ \cline{2-7} 
                                   & SVM        & 0.8858          & 0.9110          & 0.8885          & 0.8858          & 0.8858              \\ \hline
\multirow{3}{*}{MFO}               & SGD        & 0.8874          & 0.9107          & 0.8898          & 0.8874          & 0.8874              \\ \cline{2-7} 
                                   & KNN        & 0.8870          & 0.9107          & 0.8895          & 0.8870          & 0.8870              \\ \cline{2-7} 
                                   & SVM        & 0.8858          & 0.9104          & 0.8885          & 0.8858          & 0.8858              \\ \hline
\multirow{3}{*}{WOA}               & SGD        & 0.8866          & 0.9107          & 0.8892          & 0.8866          & 0.8866              \\ \cline{2-7} 
                                   & KNN        & 0.8870          & 0.9107          & 0.8895          & 0.8870          & 0.8870              \\ \cline{2-7} 
                                   & SVM        & 0.8866          & 0.9109          & 0.8892          & 0.8866          & 0.8866              \\ \hline
\multirow{3}{*}{FFA}               & SGD        & 0.8866          & 0.9100          & 0.8891          & 0.8866          & 0.8866              \\ \cline{2-7} 
                                   & KNN        & 0.8858          & 0.9094          & 0.8883          & 0.8858          & 0.8858              \\ \cline{2-7} 
                                   & SVM        & 0.8858          & 0.9102          & 0.8884          & 0.8858          & 0.8858              \\ \hline
\multirow{3}{*}{BAT}               & SGD        & 0.8858          & 0.9092          & 0.8883          & 0.8858          & 0.8858              \\ \cline{2-7} 
                                   & KNN        & 0.8850          & 0.9090          & 0.8876          & 0.8850          & 0.8850              \\ \cline{2-7} 
                                   & SVM        & 0.8850          & 0.9098          & 0.8877          & 0.8850          & 0.8850              \\ \hline
\multirow{3}{*}{HGS}               & SGD        & 0.8858          & 0.9083          & 0.8882          & 0.8858          & 0.8858              \\ \cline{2-7} 
                                   & KNN        & 0.8862          & 0.9090          & 0.8886          & 0.8862          & 0.8862              \\ \cline{2-7} 
                                   & SVM        & 0.8862          & 0.9101          & 0.8888          & 0.8862          & 0.8862              \\ \hline
\multirow{3}{*}{CGO}               & SGD        & \textbf{0.8879} & 0.9110          & \textbf{0.8995} & \textbf{0.8879} & \textbf{0.8878}     \\ \cline{2-7} 
                                   & KNN        & 0.8878          & \textbf{0.9112} & 0.8902          & 0.8878          & \textbf{0.8878}     \\ \cline{2-7} 
                                   & SVM        & 0.8866          & 0.9109          & 0.8892          & 0.8866          & 0.8866              \\ \hline
\end{tabular}

\label{tab:resultsBlood}

\end{table}

For the Blood-cell dataset, the results of the CGO method and other optimizers are shown in Table \ref{tab:resultsBlood}. The SGD, SVM, and KNN classifiers have been combined on the nine optimizers in the table. According to the table, merging the CGO algorithm with SGD surpassed other algorithms by 88.79 \%, which is the accuracy score. GWO is then used to get the same outcome as MFO (i.e., 88.74 \%). There is also 88.70 \% in the MVO. BAT and HGS had the worst score, with 88.58 \%. The CGO also had the best results on the precision metric, with 91.10 \% of the vote. Ninety-one \% (91.07 \%) was the second-best result, which belongs to MFO and WOA. Two other algorithms (BAT and HGS) performed poorly, with 90.92 \% and 90.83 \% of their respective performances, respectively. Recall results were better when using the CGO algorithm, with the best outcomes. The GWO and MFO all have the same recall (i.e., 88.74 \%). They are followed closely by the MVO 88.66 \% was reached by the FFA and WOA, whom the FFA and the WOA followed. Finally, the BAT and HGS algorithms have a worse outcome of 88.58 \%. The proposed CGO also outperformed other algorithms on F1-Score, with 89.95 \%. The MFO and GWO optimizers came in second with 88.98 \%. There are also 88.95 \% for each of the other algorithms: MVO, WOA; FFA, PSO; and BAT; correspondingly. Finally, the HGS gets the poorest performance with 88.82 \% of the population. In the CGO algorithm, 88.78 \% accuracy was attained. At the same time, MFO  was ranked second (88.74 \%) by the GWO. With 88.66 \%, WOA and FFA algorithms are next in line. Only BAT and HGS achieved a score of 88.58 \%.

\begin{figure}
    \centering
    \includegraphics
    [width=10cm, height=5cm]
    {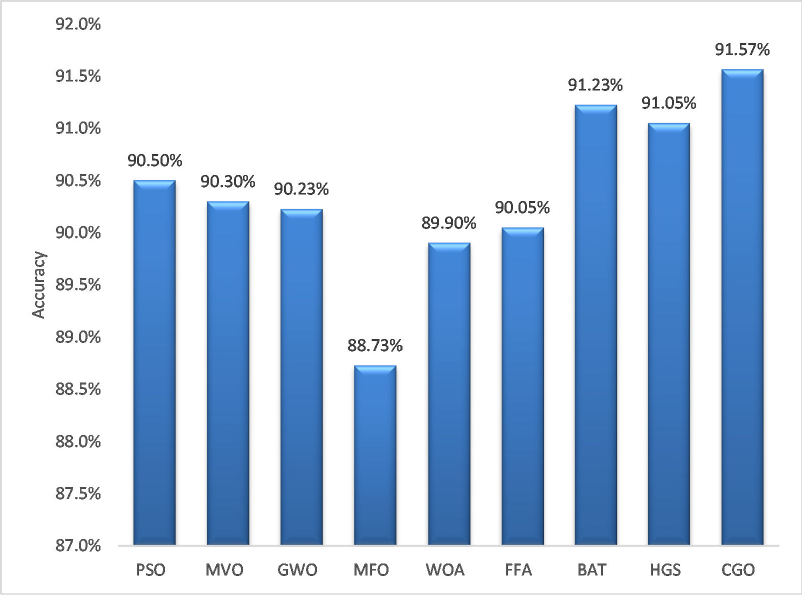}
    \caption{\label{fig:avgAcc} Average accuracy of SGD classifier of the selected datasets based on nine FS algorithms.}
    
\end{figure}

According to a different perspective, Figure \ref{fig:avgAcc} depicts the average accuracy of each feature selection optimization algorithm on the three selected datasets examined on the SGD classifier. The total average result on three databases is about 91.57 \% for the CGO, while the BAT technique comes in second with 91.23 \%. About 91.05 \% of outcomes from the HGS are better than those from the PSO. Those are followed by the MVO (90.03 \%), GWO (90.23 \%), FFA (90.05 \%), and WOA (89.90 \%). Last but not least, the MFO has the lowest performance (88.73 \%).

\begin{figure}
    \centering
    \includegraphics
    [width=10cm, height=5cm]
    {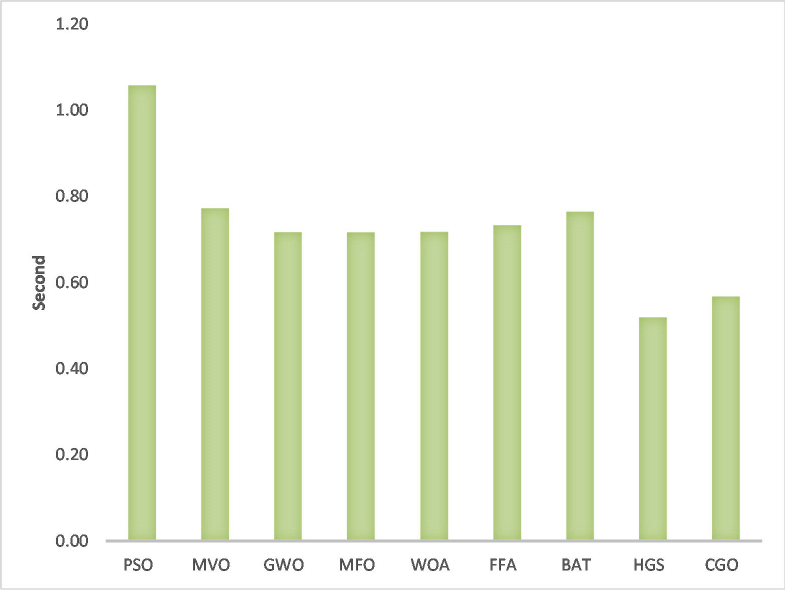}
    \caption{\label{fig:avgTime} Average execution time of nine FS methods.}
    
\end{figure}

According to a client, the complete method takes far less time to execute. Figure \ref{fig:avgTime} shows that the suggested CGO and HGS algorithms have an average execution time of 0.5672 and 0.5189 seconds for the three datasets, respectively. These results are lower than those of other algorithms that have been compared. The MFO optimizer took 0.7164 seconds to run, whereas GWO, WOA, FFA, BAT, and MVO took 0.7169 s, 0.7177 s, 0.7332 s, 0.7644 s, and 0.7723 s, respectively. The highest (or worst) execution time was attained (1.0576 s) for the PSO.

\begin{figure}
    \centering
    \includegraphics
    [width=10cm, height=5cm]
    {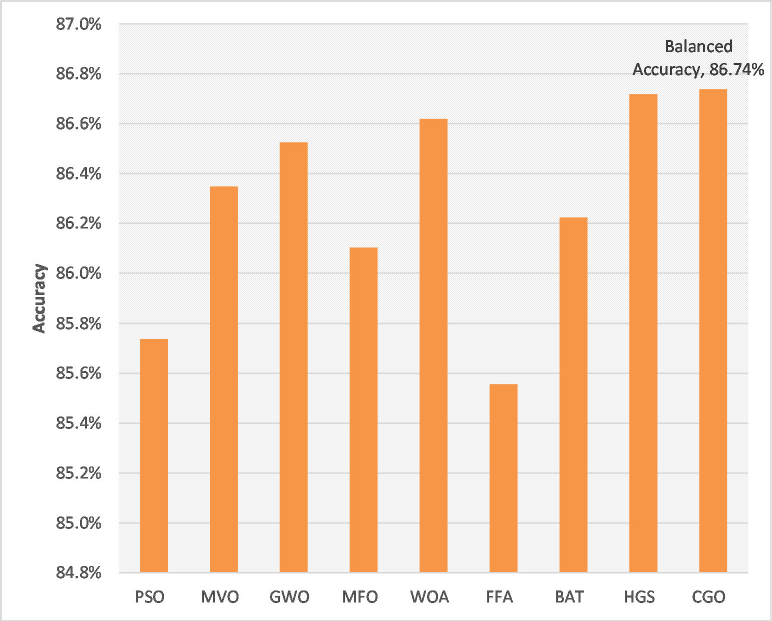}
    \caption{\label{fig:AverBA} Average balanced accuracy of the selected datasets based on nine FS algorithms.}
    
\end{figure}

Figure \ref{fig:AverBA} displays the average balanced accuracy of each feature selection approach on the three datasets, namely: ISIC-2016, PH2, and Blood-Cell, from a different perspective. On average, KNN, SVM, and SGD classifiers outperform the CGO approach by 86.74 \%; the HGS method comes in second with 86.72 \%. The WOA delivers superior results than GWO and MVO, with 86.62 and 86.53 \%, respectively. 86.22 \% of the vote goes to the BAT. After that, the MFO, PSO, and FFA optimizer obtained the lowest results, with average balanced accuracy of 86.10 \%, 85.74 \%, and 85.56 \%, respectively.

\begin{figure}
    \centering
    \includegraphics
    [width=10cm, height=5cm]
    {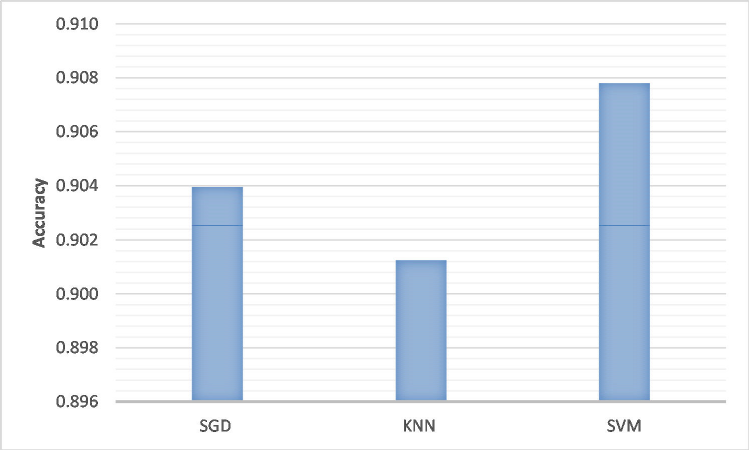}
    \caption{\label{fig:avgAccCls} The averaged results of the selected dataset  in terms of accuracy metric using the three classifiers.}
    
\end{figure}

The SGD, SVM, and KNN classifier's average accuracy on the three selected datasets were shown in Figure \ref{fig:avgAccCls} on various techniques for optimization (i.e., the nine optimizers, which introduced before). In the figure, we can see that the SVM outperformed other classifiers on the accuracy metric. To be more specific, the SVM achieved 90.78 \% accuracy, whereas the KNN achieved 90.13 \% accuracy. In the end, the SGD algorithm achieved 90.40 \%.

\begin{figure}
    \centering
    \includegraphics
    [width=10cm, height=5cm]
    {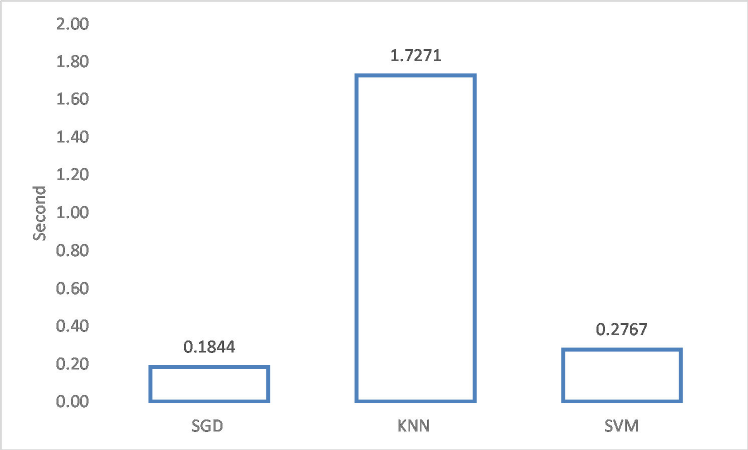}
    \caption{\label{fig:avgTimeCls} Average execution time of the three classifiers.}
    
\end{figure}

However, the time to complete the full procedure is shorter than for a user. As a result, the average execution time of the optimization algorithms for the three databases is presented in Figure \ref{fig:avgTimeCls}. SGD's classification algorithm took the least amount of time, according to the results. Then comes the SVM classifier, which takes 0.2767 seconds to complete its task. 1.7271 seconds is the longest (and therefore the worst) time for another classifier, KNN.

To sum it up, the CGO optimization technique paired with the SGD classifier earned the greatest accuracy metric among all combinations for the ISIC-2016, PH2, and Blood datasets. Moreover, the SGD outperforms other classification algorithms (i.e., KNN and SVM) according to the results.


\subsubsection{Comparison with the literature studies}


This section compares with other state-of-the-art medical image classification techniques. Table \ref{tab:resultsSOTA} shows the results of state-of-the-art methods. The development of high-accuracy technology for medical image classification is a major undertaking. It is important to compare our strategy to other models that have been tested on the same datasets. Using ISIC-2016, PH2, and Blood-Cell datasets, Table \ref{tab:resultsSOTA} evaluates the performance of several techniques for disease identification.

For the ISIC-2016 dataset, the following advanced skin cancer identification methods were compared: Based on segregation and then validation \cite{yu2016automated}; relied on feature-fusion \cite{ge2017exploiting}; correlated with fisher-coding and deep residual networks \cite{yu2018melanoma}; multi-CNN interactive learning model \cite{zhang2019medical}; ensemble method \cite{pathan2019automated}; and integrating fisher-vector and CNN fusion \cite{yu2020convolutional}. To differentiate characteristics, a fine-grained classification concept is applied \cite{wei2020automatic}. 

For the PH2 dataset, the following sophisticated technologies for melanoma diagnosis were evaluated with each other Involves an artificial neural network, as introduced in \cite{ozkan2017skin}, in addition, they developed a decision-aid system. It was proposed by \cite{moradi2019kernel} to use sparse kernel models to represent feature data in a high-dimensional feature vector. According to \cite{al2020automatic}, U-Net can be used to detect malignant tumors automatically. As part of their IoT system, \cite{rodrigues2020new} employed transfer learning and CNN. A hierarchical architecture founded on two-dimensional pixels in the image and ResNet was introduced in  \cite{afza2021hierarchical} for advanced DL.

As a result of the CNN solution, the SVM-based classifiers were able to classify data, as proposed in \cite{habibzadeh2013white}. Besides, a granularity feature and SVM are used in \cite{zhao2017automatic}. In order to identify and count essential blood cells in the Blood-Cell dataset, they used the following identifying and counting methods. 
In order to automate the entire procedure, CNNs were presented as a DL method in \cite{sharma2019white}.

\begin{figure}
    \centering
    \includegraphics[width=15cm]{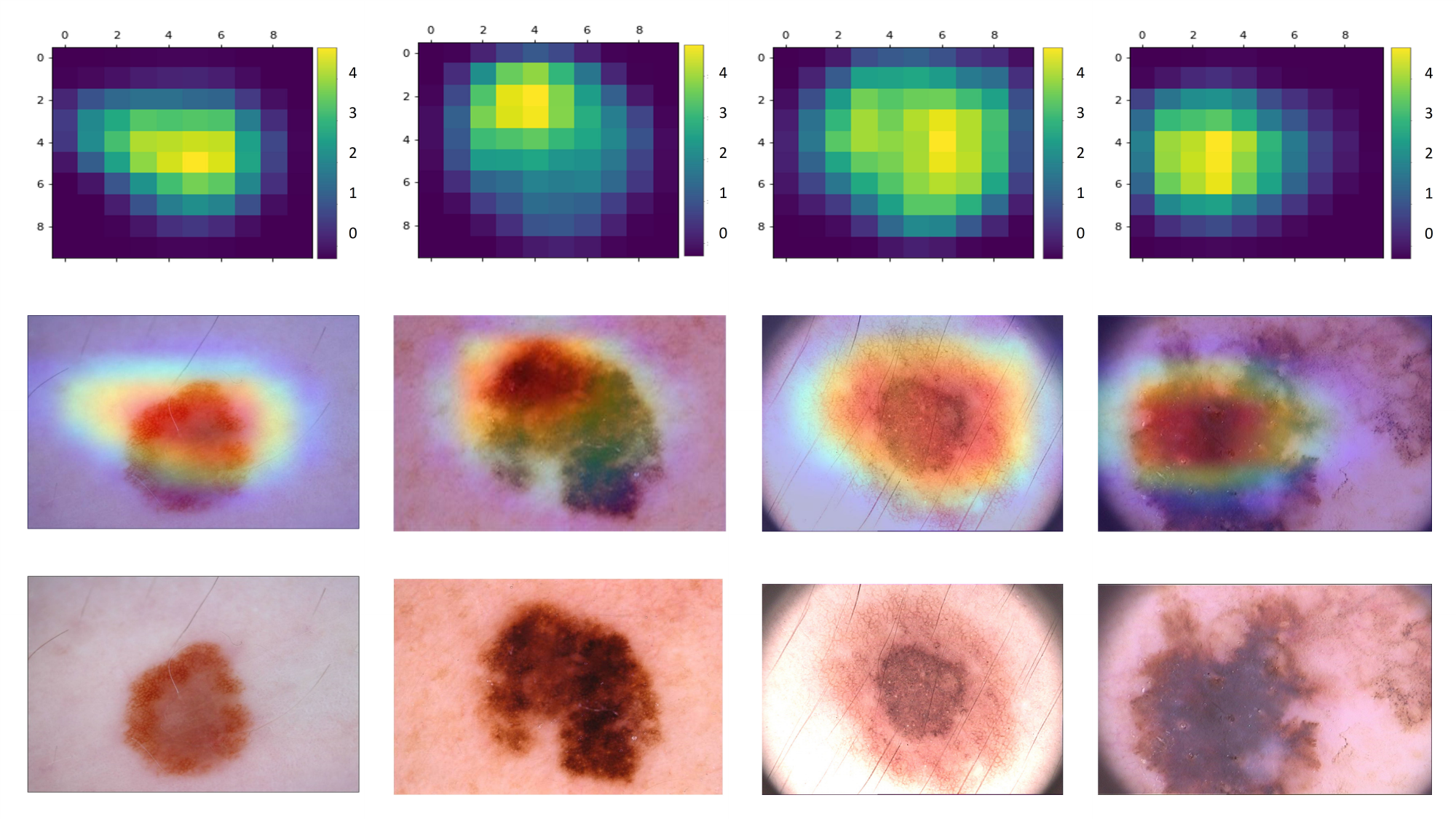}
    \caption{\label{fig:Grad} Grad-CAM heatmaps on the skin images using MobileNetV3 model.}
\end{figure}

\begin{table}
\centering
\caption{\label{tab:resultsSOTA} Accuracy results (\%) of the existing approaches.}

\begin{tabular}{|c|c|c|c|c|}
\hline
Source & Dataset & Year & Classification Model & Accuracy (\%)  \\ \hline

\cite{yu2016automated}& ISIC-2016 &2016&	CUMED&	85.50 
\\ \hline

\cite{ge2017exploiting}& ISIC-2016 &2017	&BL-CNN&	85.00
\\ \hline

\cite{yu2018melanoma}& ISIC-2016 &2018&	DCNN-FV&86.81
\\ \hline

\cite{zhang2019medical}& ISIC-2016 &2019&	MC-CNN&	86.30 
\\ \hline

\cite{pathan2019automated}& ISIC-2016 &2019	&KNORA-E&	88.00
\\ \hline

\cite{yu2020convolutional}& ISIC-2016 &2020	&MFA&86.81	
 \\ \hline

\cite{wei2020automatic}& ISIC-2016 &2020&	FUSION	&87.60
\\ \hline


\textbf{Our} & \textbf{ISIC-2016} &\textbf{present}	& \textbf{CGO + SGD} 	& \textbf{88.39} \\ \hline

\cite{ozkan2017skin} & PH2 & 2017 & ANN  &  92.50 
  \\ \hline
\cite{moradi2019kernel} & PH2& 2019 & Kernel Sparse &    93.50 
    \\ \hline
\cite{al2020automatic} &PH2& 2020 & DenseNet201 + SVM & 92.00 
 \\ \hline
\cite{rodrigues2020new}&PH2&2020	&DenseNet201 + KNN	&	93.16
\\ \hline
\cite{afza2021hierarchical}&PH2&2021&	ResNet50 + NB	&95.40
\\ \hline


\textbf{Our} &\textbf{PH2} &\textbf{present}	& \textbf{CGO + SGD} 	& \textbf{97.52} \\ \hline

\cite{habibzadeh2013white} &Blood-Cell& 2013 & CNN + SVM  &  85.00 
  \\ \hline
\cite{zhao2017automatic} &Blood-Cell& 2017 & CNN & 87.08 
 \\ \hline
 
\cite{sharma2019white} &Blood-Cell& 2019 & CNN + Augmentation &    87.00 
    \\ \hline


\textbf{Our} & \textbf{Blood-Cell} & \textbf{present}	& \textbf{CGO + SGD} 	& \textbf{88.79} \\ \hline

\end{tabular}

\end{table}

\color{black}

\section{Discussion} \label{dis}

\color{black}
The bottom line is that we can remove superfluous features from high-dimensional medical image representations obtained by CNN (i.e., MobileNetV3). The MobileNetV3 models achieved the effective performance as feature extractor in our work. The class activation map for the MobileNetV3 model was prepared where the activation provided by the last layer is represented as an overlayed heat map, as shown in Figure \ref{fig:Grad}. In the figure, the red regions represent the most important discriminative regions, while the other colored regions are less important.

\begin{figure}
    \centering
    \includegraphics
    [width=13cm,height=7cm]
    {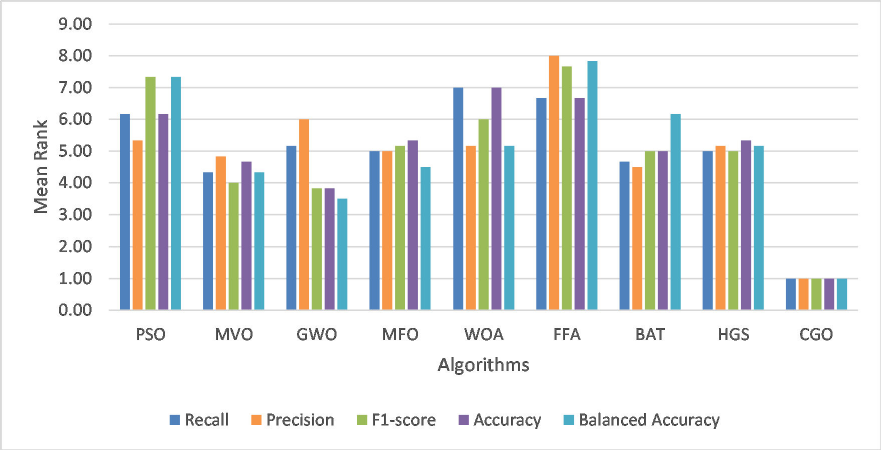}
    \caption{\label{fig:MeanRank} The mean rank of FD test on several feature selection algorithms on SGD classifier.}
    
\end{figure}

In order to include a more complex comparison among different algorithms, we have used the Friedman (FD) test. The FD test is nonparametric that calculates and ranks the statistical value. In \cite{derrac2011practical},  The FD test is used to determine whether there is a significant difference between different methods. Furthermore, Figure \ref{fig:MeanRank} compares the CGO method to the nine optimization techniques on the three datasets in terms of Recall, Precision, F1-measure, Accuracy, and Balanced Accuracy. When the CGO's results are analyzed using the five metrics, it is clear that the CGO algorithm surpasses the others.
In terms of balanced accuracy, the CGO has the lowest mean ranking of 1, following the GWO has the mean rank of 3.50. MFO and MVO have nearly identical mean levels, with 4. WOA and HGS have a mean rank of 4.17. Finally, BAT, PSO, and FFA are lower than the others, with a mean rank of 6.17, 7.33, and 7.83, respectively.
According to the FD test results for accuracy, CGO is also better than others, with a mean rank of 1. They were followed by GWO, which achieved 3.83. On the other hand, MVO and BAT have the same mean level of 5, whereas MFO and HGS have 5.33. Lastly, PSO, FFA, and WOA have the highest mean ranking.
Furthermore, we discovered that the CGO in the F1-score measure has the best mean rank of 1, and the GWO and MVO have the second and third mean ranks of 3.83 and 4, respectively. BAT and HGS have nearly identical mean levels (i.e., 5). Finally, MFO and WOA have a mean rank of 5.17 and 6.00, respectively.
Finally, PSO and FFA have lower than the others, with a mean rank of 7.33 and 7.67, respectively.
Finally, the precision measure difference between the CGO and the BAT, MVO, MFO, WOA, HGS, PSO, GWO, and FFA optimization algorithms averages 4.5, 4.83, 5, 5.17, 5.17, 5.33, 6, and 8, respectively.
According to the FD test results for recall, CGO is also better than others, with a mean rank of 1. They were followed by MVO, which achieved 4.33. BAT has a mean rank level of 4.67, whereas MFO and HGS have 5. Lastly, GWO, PSO, FFA, and WOA have the highest mean ranking. 
As a result of Friedman's test, there is a noticeable difference between the proposed model and the other models (where the p-value is less than 0.05), as shown in Figure \ref{fig:MeanRank}.

This reasons support that our approach obtains the best results. Thus, CGO is an effective search algorithm for tackling complex optimization issues; therefore, it is critical to pick its parameters carefully. For example, when the clusters of the population in CGO were analyzed, CGO worked better when the population of an optimal solution was classified into two parts. Second, CGO performed searches more consistently than other methods, as evidenced by lower standard deviation values in the results. Finally, CGO's exploration and exploitation techniques are successfully applied since they worked equally well on datasets with a wide variety of dimensions, making FS challenges adaptable.

However, our approach also has some limitations, mainly in time and memory complexity. Therefore, we are currently working on trying to improve the efficiency of our approach. In fact, we are assessing to take into account other augmentation procedures, as introduced in \cite{oyewola2022novel}. Moreover, we plan to use other deep learning models such as swin or vision transformer, which achieved the best results and have been more recently used in different computer vision tasks.
\color{black}

\section{Conclusion and future work} \label{c}

\color{black}
The automatic medical image classification task has been expanding rapidly in recent years. However, existing approaches are still incapable of achieving good performance due to the similarity in physical attributes of images, the diversity of medical experience, and a small medical image dataset.
\color{black}
Therefore, this paper demonstrates a new method of classifying medical images that uses the IoMT system to help clinicians and patients make a quick and advanced diagnosis of diseases in any area.
The proposed system relies on the classification models trained in the cloud center before being used, after extracting features from the medical images acquired from IoT nodes and passing them to fog computing. To obtain the features, MobileNetV3 was used. The MobileNetV3 was fine-tuned on medical image datasets to generate higher sophisticated and informative representations and retrieve feature vector representation. 
\color{black}
After that, we used a new metaheuristic method in binary form (as Chaos Game Optimization) to reduce the features representation space.
This algorithm leads to an enhancement for the convergence rate towards the optimal subset of relevant features. Therefore, 
\color{black}
CGO produces a high convergence speed. This indicates it avoids trapping in local optima. Thus, it successfully balances the exploration and exploitation phases because of the fast determination of the threshold values and the high accuracy presented in the results.
\color{black}
The learned model's efficiency is evaluated either by transmitting it to a tested medical images cloud center or by using fog computing with a clone of the learning algorithm. 
Our experiments were applied on three databases, ISIC-2016, PH2, and Blood-Cell. 
\color{black}
According to the results, the new CGO optimization method outperforms other existing feature selection methods. The work evaluated the combinations of nine optimizers with three different classifier configurations. The most significant results for accuracy, F1-Score, recall, and precision metrics for these datasets were achieved with the CGO optimizer combined with the SGD classifier. For ISIC-2016, the accuracy value was 88.39 \%, for PH2 the accuracy was 97.52\% and, finally, for Blood-Cell the accuracy was 88.79\%. 
\color{black}
Furthermore, the results of the comparisons with some other state-of-the-art medical image classification technologies demonstrated that the created IoMT methodology is an appropriate mechanism. In the near future, \color{black}
this system would be available in hospitals with the aim of monitoring the patients' condition from home.  Patients would automatically send a report to the hospital through the connected devices, with vital information about blood pressure, insulin levels, etc. Then, professional staff at the hospital would follow up each case and, if needed, would directly communicate to the patient.
\color{black}

\color{black}
However, there are still some limitations to the proposed model, being the most relevant the requirements for computational resources, that is, more time is needed to obtain the results, and also the requirements for memory resources. 
\color{black}
We are currently working on lowering complexity and enhancing the efficiency of the suggested system. Also, we plan to propose a CGO based multi-objective feature selection approach for high dimensional data with a small instance to simultaneously maximize the classification performance and minimize the number of features, using more efficient classifiers. Additionally, automatic cluster number determination and the application of hyper-heuristic approaches in FS can also be an exciting line of research. Moreover, a more comprehensive volume of medical data will be evaluated in the future study. Finally, merging several classifications algorithms is an attractive object of investigation that could allow practitioners to influence the performance of existing methods.
\color{black}

\section*{Acknowledgement}
This work has received financial support from the European Regional Development Fund (ERDF) and the Galician Regional Government, under the agreement for funding the Atlantic Research Center for Information and Communication Technologies (atlanTTic). This work was also supported by the Spanish Government under re-search project “Enhancing Communication Protocols with Machine Learning while Protecting Sensitive Data (COMPROMISE)" (PID2020-113795RB-C33/AEI/10.13039/501100011033).


\color{black}
\section*{Abbreviations}

The following abbreviations are used in this manuscript:

\begin{table}[H]
\centering
\begin{tabular}{ll}
\hline
\textbf{Abbreviations} & \textbf{Definition} \\ \hline
IoMT & Internet of Medical Things \\
TL & Transfer Learning \\
IoT & Internet of Things \\
DL & Deep learning \\
CNN & Convolution Neural Networks \\
FS & Feature Selection \\
CGO & Chaos Game Optimization \\
PSO & Particle Swarm Optimization \\
GWO & Grey Wolf Optimization \\
GA & Genetic Algorithm \\
MVO & Multi-Verse Optimizer \\
MFO & Moth-Flame Optimization \\
BAT & Bat Algorithm \\
HGS & Hunger Games Search \\
FFA & Firefly Algorithms \\
WOA & Whale Optimization Algorithm \\
NAS & network architecture search \\
MLP & multilayer perceptron \\
SE & Squeeze-And-Excite block \\
ReLU & Rectified linear unit \\
Conv & point-wise convolution \\
BN & batch normalization \\
FC & fully connected layers \\
SGD & stochastic gradient descent \\
KNN & k-Nearest Neighbor \\
SVM & Support Vector Machines \\
MRI & magnetic resonance imaging \\
CAD & Computed Aided Diagnosis \\
ROI & Region of interest \\
FD & Friedman test \\ \hline
\end{tabular}
\end{table}

\color{black}

\bibliographystyle{elsarticle-num}
\bibliography{sample}

\begin{thebibliography}{10}
\expandafter\ifx\csname url\endcsname\relax
  \def\url#1{\texttt{#1}}\fi
\expandafter\ifx\csname urlprefix\endcsname\relax\def\urlprefix{URL }\fi
\expandafter\ifx\csname href\endcsname\relax
  \def\href#1#2{#2} \def\path#1{#1}\fi

\bibitem{ren2022application}
W.~Ren, X.~Wu, Application of intelligent medical equipment management system based on internet of things technology, Journal of Healthcare Engineering 2022 (2022).

\bibitem{alharbi2021botnet}
A.~Alharbi, W.~Alosaimi, H.~Alyami, H.~T. Rauf, R.~Dama{\v{s}}evi{\v{c}}ius, Botnet attack detection using local global best bat algorithm for industrial internet of things, Electronics 10~(11) (2021) 1341.

\bibitem{mehra2022lbecr}
P.~S. Mehra, Lbecr: load balanced, efficient clustering and routing protocol for sustainable internet of things in smart cities, Journal of Ambient Intelligence and Humanized Computing (2022) 1--23.

\bibitem{sinha2022recent}
B.~B. Sinha, R.~Dhanalakshmi, Recent advancements and challenges of internet of things in smart agriculture: A survey, Future Generation Computer Systems 126 (2022) 169--184.

\bibitem{tamulis2022affective}
{\v{Z}}.~Tamulis, M.~Vasiljevas, R.~Dama{\v{s}}evi{\v{c}}ius, R.~Maskeliunas, S.~Misra, Affective computing for ehealth using low-cost remote internet of things-based emg platform, in: Intelligent Internet of Things for Healthcare and Industry, Springer, 2022, pp. 67--81.

\bibitem{abd2022medical}
M.~Abd~Elaziz, A.~Mabrouk, A.~Dahou, S.~A. Chelloug, Medical image classification utilizing ensemble learning and levy flight-based honey badger algorithm on 6g-enabled internet of things, Computational Intelligence and Neuroscience 2022 (2022).

\bibitem{hossen2022federated}
M.~N. Hossen, V.~Panneerselvam, D.~Koundal, K.~Ahmed, F.~M. Bui, S.~M. Ibrahim, Federated machine learning for detection of skin diseases and enhancement of internet of medical things (iomt) security, IEEE journal of biomedical and health informatics (2022).

\bibitem{karar2022intelligent}
M.~E. Karar, B.~Alotaibi, M.~Alotaibi, Intelligent medical iot-enabled automated microscopic image diagnosis of acute blood cancers, Sensors 22~(6) (2022) 2348.

\bibitem{maqsood2022ttcnn}
S.~Maqsood, R.~Dama{\v{s}}evi{\v{c}}ius, R.~Maskeli{\=u}nas, Ttcnn: A breast cancer detection and classification towards computer-aided diagnosis using digital mammography in early stages, Applied Sciences 12~(7) (2022) 3273.

\bibitem{mabrouk2022pneumonia}
A.~Mabrouk, R.~P. D{\'\i}az~Redondo, A.~Dahou, M.~Abd~Elaziz, M.~Kayed, Pneumonia detection on chest x-ray images using ensemble of deep convolutional neural networks, Applied Sciences 12~(13) (2022) 6448.

\bibitem{odusami2022intelligent}
M.~Odusami, R.~Maskeli{\=u}nas, R.~Dama{\v{s}}evi{\v{c}}ius, An intelligent system for early recognition of alzheimer’s disease using neuroimaging, Sensors 22~(3) (2022) 740.

\bibitem{ogundokun2022medical}
R.~O. Ogundokun, S.~Misra, M.~Douglas, R.~Dama{\v{s}}evi{\v{c}}ius, R.~Maskeli{\=u}nas, Medical internet-of-things based breast cancer diagnosis using hyperparameter-optimized neural networks, Future Internet 14~(5) (2022) 153.

\bibitem{taryudi2021smart}
T.~Taryudi, L.~Lindayani, H.~Purnama, A.~Mutiar, Smart bed notification system base on internet of things for fall prevention in patients with stroke, Journal of Medical Engineering \& Technology (2021) 1--6.

\bibitem{budd2021survey}
S.~Budd, E.~C. Robinson, B.~Kainz, A survey on active learning and human-in-the-loop deep learning for medical image analysis, Medical Image Analysis 71 (2021) 102062.

\bibitem{mabrouk2021seopinion}
A.~Mabrouk, R.~P.~D. Redondo, M.~Kayed, Seopinion: Summarization and exploration of opinion from e-commerce websites, Sensors 21~(2) (2021) 636.

\bibitem{parsons2021automatic}
Z.~Parsons, S.~Banitaan, Automatic identification of chagas disease vectors using data mining and deep learning techniques, Ecological Informatics 62 (2021) 101270.

\bibitem{mabrouk2020deep}
A.~Mabrouk, R.~P.~D. Redondo, M.~Kayed, Deep learning-based sentiment classification: A comparative survey, IEEE Access 8 (2020) 85616--85638.

\bibitem{zoetmulder2022domain}
R.~Zoetmulder, E.~Gavves, M.~Caan, H.~Marquering, Domain-and task-specific transfer learning for medical segmentation tasks, Computer Methods and Programs in Biomedicine 214 (2022) 106539.

\bibitem{nguyen2022tatl}
D.~M. Nguyen, T.~T. Nguyen, H.~Vu, Q.~Pham, M.-D. Nguyen, B.~T. Nguyen, D.~Sonntag, Tatl: task agnostic transfer learning for skin attributes detection, Medical Image Analysis (2022) 102359.

\bibitem{liu2022margin}
Z.~Liu, Z.~Zhu, S.~Zheng, Y.~Liu, J.~Zhou, Y.~Zhao, Margin preserving self-paced contrastive learning towards domain adaptation for medical image segmentation, IEEE Journal of Biomedical and Health Informatics (2022).

\bibitem{ren2022robustness}
S.~Ren, C.~Q. Li, Robustness of transfer learning to image degradation, Expert Systems with Applications 187 (2022) 115877.

\bibitem{niu2021distant}
S.~Niu, M.~Liu, Y.~Liu, J.~Wang, H.~Song, Distant domain transfer learning for medical imaging, IEEE Journal of Biomedical and Health Informatics (2021).

\bibitem{adel2022improving}
H.~Adel, A.~Dahou, A.~Mabrouk, M.~Abd~Elaziz, M.~Kayed, I.~M. El-Henawy, S.~Alshathri, A.~Amin~Ali, Improving crisis events detection using distilbert with hunger games search algorithm, Mathematics 10~(3) (2022) 447.

\bibitem{manimurugan2022two}
S.~Manimurugan, S.~Almutairi, M.~M. Aborokbah, C.~Narmatha, S.~Ganesan, N.~Chilamkurti, R.~A. Alzaheb, H.~Almoamari, Two-stage classification model for the prediction of heart disease using iomt and artificial intelligence, Sensors 22~(2) (2022) 476.

\bibitem{khelili2022iomt}
M.~Khelili, S.~Slatnia, O.~Kazar, S.~Harous, Iomt-fog-cloud based architecture for covid-19 detection, Biomedical Signal Processing and Control 76 (2022) 103715.

\bibitem{khalil2022efficient}
A.~A. Khalil, F.~E~Ibrahim, M.~Y. Abbass, N.~Haggag, Y.~Mahrous, A.~Sedik, Z.~Elsherbeeny, A.~A. Khalaf, M.~Rihan, W.~El-Shafai, et~al., Efficient anomaly detection from medical signals and images with convolutional neural networks for internet of medical things (iomt) systems, International Journal for Numerical Methods in Biomedical Engineering 38~(1) (2022) e3530.

\bibitem{kumar2022depress}
A.~Kumar, S.~R. Sangwan, A.~Arora, V.~G. Menon, Depress-dcnf: A deep convolutional neuro-fuzzy model for detection of depression episodes using iomt, Applied Soft Computing 122 (2022) 108863.

\bibitem{han2020internet}
T.~Han, V.~X. Nunes, L.~F. D.~F. Souza, A.~G. Marques, I.~C.~L. Silva, M.~A. A.~F. Junior, J.~Sun, P.~P. Reboucas~Filho, Internet of medical things—based on deep learning techniques for segmentation of lung and stroke regions in ct scans, IEEE Access 8 (2020) 71117--71135.

\bibitem{cheplygina2019not}
V.~Cheplygina, M.~de~Bruijne, J.~P. Pluim, Not-so-supervised: a survey of semi-supervised, multi-instance, and transfer learning in medical image analysis, Medical image analysis 54 (2019) 280--296.

\bibitem{ayan2019diagnosis}
E.~Ayan, H.~M. {\"U}nver, Diagnosis of pneumonia from chest x-ray images using deep learning, in: 2019 Scientific Meeting on Electrical-Electronics \& Biomedical Engineering and Computer Science (EBBT), Ieee, 2019, pp. 1--5.

\bibitem{chouhan2020novel}
V.~Chouhan, S.~K. Singh, A.~Khamparia, D.~Gupta, P.~Tiwari, C.~Moreira, R.~Dama{\v{s}}evi{\v{c}}ius, V.~H.~C. De~Albuquerque, A novel transfer learning based approach for pneumonia detection in chest x-ray images, Applied Sciences 10~(2) (2020) 559.

\bibitem{zhang2019medical}
J.~Zhang, Y.~Xie, Q.~Wu, Y.~Xia, Medical image classification using synergic deep learning, Medical image analysis 54 (2019) 10--19.

\bibitem{pathan2019automated}
S.~Pathan, K.~G. Prabhu, P.~Siddalingaswamy, Automated detection of melanocytes related pigmented skin lesions: A clinical framework, Biomedical Signal Processing and Control 51 (2019) 59--72.

\bibitem{yu2020convolutional}
Z.~Yu, F.~Jiang, F.~Zhou, X.~He, D.~Ni, S.~Chen, T.~Wang, B.~Lei, Convolutional descriptors aggregation via cross-net for skin lesion recognition, Applied Soft Computing 92 (2020) 106281.

\bibitem{wei2020automatic}
L.~Wei, K.~Ding, H.~Hu, Automatic skin cancer detection in dermoscopy images based on ensemble lightweight deep learning network, IEEE Access 8 (2020) 99633--99647.

\bibitem{samala2018evolutionary}
R.~K. Samala, H.-P. Chan, L.~M. Hadjiiski, M.~A. Helvie, C.~Richter, K.~Cha, Evolutionary pruning of transfer learned deep convolutional neural network for breast cancer diagnosis in digital breast tomosynthesis, Physics in Medicine \& Biology 63~(9) (2018) 095005.

\bibitem{da2018convolutional}
G.~L.~F. da~Silva, T.~L.~A. Valente, A.~C. Silva, A.~C. de~Paiva, M.~Gattass, Convolutional neural network-based pso for lung nodule false positive reduction on ct images, Computer methods and programs in biomedicine 162 (2018) 109--118.

\bibitem{vijh2020brain}
S.~Vijh, S.~Sharma, P.~Gaurav, Brain tumor segmentation using otsu embedded adaptive particle swarm optimization method and convolutional neural network, in: Data visualization and knowledge engineering, Springer, 2020, pp. 171--194.

\bibitem{shankar2019alzheimer}
K.~Shankar, S.~Lakshmanaprabu, A.~Khanna, S.~Tanwar, J.~J. Rodrigues, N.~R. Roy, Alzheimer detection using group grey wolf optimization based features with convolutional classifier, Computers \& Electrical Engineering 77 (2019) 230--243.

\bibitem{goel2021optconet}
T.~Goel, R.~Murugan, S.~Mirjalili, D.~K. Chakrabartty, Optconet: an optimized convolutional neural network for an automatic diagnosis of covid-19, Applied Intelligence 51~(3) (2021) 1351--1366.

\bibitem{elhoseny2019optimal}
M.~Elhoseny, K.~Shankar, Optimal bilateral filter and convolutional neural network based denoising method of medical image measurements, Measurement 143 (2019) 125--135.

\bibitem{zhang2020skin}
N.~Zhang, Y.-X. Cai, Y.-Y. Wang, Y.-T. Tian, X.-L. Wang, B.~Badami, Skin cancer diagnosis based on optimized convolutional neural network, Artificial intelligence in medicine 102 (2020) 101756.

\bibitem{el2021clustering}
E.~El-Shafeiy, K.~M. Sallam, R.~K. Chakrabortty, A.~A. Abohany, A clustering based swarm intelligence optimization technique for the internet of medical things, Expert Systems with Applications 173 (2021) 114648.

\bibitem{gutman2016skin}
D.~Gutman, N.~C. Codella, E.~Celebi, B.~Helba, M.~Marchetti, N.~Mishra, A.~Halpern, Skin lesion analysis toward melanoma detection: A challenge at the international symposium on biomedical imaging (isbi) 2016, hosted by the international skin imaging collaboration (isic), arXiv preprint arXiv:1605.01397 (2016).

\bibitem{mendoncca2013ph}
T.~Mendon{\c{c}}a, P.~M. Ferreira, J.~S. Marques, A.~R. Marcal, J.~Rozeira, Ph 2-a dermoscopic image database for research and benchmarking, in: 2013 35th annual international conference of the IEEE engineering in medicine and biology society (EMBC), IEEE, 2013, pp. 5437--5440.

\bibitem{liang2018combining}
G.~Liang, H.~Hong, W.~Xie, L.~Zheng, Combining convolutional neural network with recursive neural network for blood cell image classification, IEEE Access 6 (2018) 36188--36197.

\bibitem{tran2019video}
D.~Tran, H.~Wang, L.~Torresani, M.~Feiszli, Video classification with channel-separated convolutional networks, in: Proceedings of the IEEE/CVF International Conference on Computer Vision, 2019, pp. 5552--5561.

\bibitem{ji2020action}
J.~Ji, R.~Krishna, L.~Fei-Fei, J.~C. Niebles, Action genome: Actions as compositions of spatio-temporal scene graphs, in: Proceedings of the IEEE/CVF Conference on Computer Vision and Pattern Recognition, 2020, pp. 10236--10247.

\bibitem{liu2021ntire}
J.~Liu, N.~Inkawhich, O.~Nina, R.~Timofte, Ntire 2021 multi-modal aerial view object classification challenge, in: Proceedings of the IEEE/CVF Conference on Computer Vision and Pattern Recognition, 2021, pp. 588--595.

\bibitem{ignatov2021real}
A.~Ignatov, A.~Romero, H.~Kim, R.~Timofte, Real-time video super-resolution on smartphones with deep learning, mobile ai 2021 challenge: Report, in: Proceedings of the IEEE/CVF Conference on Computer Vision and Pattern Recognition, 2021, pp. 2535--2544.

\bibitem{howard2017mobilenets}
A.~G. Howard, M.~Zhu, B.~Chen, D.~Kalenichenko, W.~Wang, T.~Weyand, M.~Andreetto, H.~Adam, Mobilenets: Efficient convolutional neural networks for mobile vision applications, arXiv preprint arXiv:1704.04861 (2017).

\bibitem{howard2019searching}
A.~Howard, M.~Sandler, G.~Chu, L.-C. Chen, B.~Chen, M.~Tan, W.~Wang, Y.~Zhu, R.~Pang, V.~Vasudevan, et~al., Searching for mobilenetv3, in: Proceedings of the IEEE/CVF International Conference on Computer Vision, 2019, pp. 1314--1324.

\bibitem{zhang2018shufflenet}
X.~Zhang, X.~Zhou, M.~Lin, J.~Sun, Shufflenet: An extremely efficient convolutional neural network for mobile devices, in: Proceedings of the IEEE conference on computer vision and pattern recognition, 2018, pp. 6848--6856.

\bibitem{zoph2018learning}
B.~Zoph, V.~Vasudevan, J.~Shlens, Q.~V. Le, Learning transferable architectures for scalable image recognition, in: Proceedings of the IEEE conference on computer vision and pattern recognition, 2018, pp. 8697--8710.

\bibitem{tan2019mnasnet}
M.~Tan, B.~Chen, R.~Pang, V.~Vasudevan, M.~Sandler, A.~Howard, Q.~V. Le, Mnasnet: Platform-aware neural architecture search for mobile, in: Proceedings of the IEEE/CVF Conference on Computer Vision and Pattern Recognition, 2019, pp. 2820--2828.

\bibitem{tan2019efficientnet}
M.~Tan, Q.~Le, Efficientnet: Rethinking model scaling for convolutional neural networks, in: International Conference on Machine Learning, PMLR, 2019, pp. 6105--6114.

\bibitem{he2016deep}
K.~He, X.~Zhang, S.~Ren, J.~Sun, Deep residual learning for image recognition, in: Proceedings of the IEEE conference on computer vision and pattern recognition, 2016, pp. 770--778.

\bibitem{ramachandran2017searching}
P.~Ramachandran, B.~Zoph, Q.~V. Le, Searching for activation functions, arXiv preprint arXiv:1710.05941 (2017).

\bibitem{elfwing2018sigmoid}
S.~Elfwing, E.~Uchibe, K.~Doya, Sigmoid-weighted linear units for neural network function approximation in reinforcement learning, Neural Networks 107 (2018) 3--11.

\bibitem{talatahari2021chaos}
S.~Talatahari, M.~Azizi, Chaos game optimization: a novel metaheuristic algorithm, Artificial Intelligence Review 54~(2) (2021) 917--1004.

\bibitem{liu2019bad}
S.~Liu, D.~Papailiopoulos, D.~Achlioptas, Bad global minima exist and sgd can reach them, arXiv preprint arXiv:1906.02613 (2019).

\bibitem{eberhart1995new}
R.~Eberhart, J.~Kennedy, A new optimizer using particle swarm theory, in: MHS'95. Proceedings of the Sixth International Symposium on Micro Machine and Human Science, Ieee, 1995, pp. 39--43.

\bibitem{mirjalili2016multi}
S.~Mirjalili, S.~M. Mirjalili, A.~Hatamlou, Multi-verse optimizer: a nature-inspired algorithm for global optimization, Neural Computing and Applications 27~(2) (2016) 495--513.

\bibitem{emary2016binary}
E.~Emary, H.~M. Zawbaa, A.~E. Hassanien, Binary grey wolf optimization approaches for feature selection, Neurocomputing 172 (2016) 371--381.

\bibitem{mirjalili2015moth}
S.~Mirjalili, Moth-flame optimization algorithm: A novel nature-inspired heuristic paradigm, Knowledge-based systems 89 (2015) 228--249.

\bibitem{mirjalili2016whale}
S.~Mirjalili, A.~Lewis, The whale optimization algorithm, Advances in engineering software 95 (2016) 51--67.

\bibitem{yang2010firefly}
X.-S. Yang, Firefly algorithm, levy flights and global optimization, in: Research and development in intelligent systems XXVI, Springer, 2010, pp. 209--218.

\bibitem{yang2010new}
X.-S. Yang, A new metaheuristic bat-inspired algorithm, in: Nature inspired cooperative strategies for optimization (NICSO 2010), Springer, 2010, pp. 65--74.

\bibitem{yang2021hunger}
Y.~Yang, H.~Chen, A.~A. Heidari, A.~H. Gandomi, Hunger games search: Visions, conception, implementation, deep analysis, perspectives, and towards performance shifts, Expert Systems with Applications 177 (2021) 114864.

\bibitem{yu2016automated}
L.~Yu, H.~Chen, Q.~Dou, J.~Qin, P.-A. Heng, Automated melanoma recognition in dermoscopy images via very deep residual networks, IEEE transactions on medical imaging 36~(4) (2016) 994--1004.

\bibitem{ge2017exploiting}
Z.~Ge, S.~Demyanov, B.~Bozorgtabar, M.~Abedini, R.~Chakravorty, A.~Bowling, R.~Garnavi, Exploiting local and generic features for accurate skin lesions classification using clinical and dermoscopy imaging, in: 2017 IEEE 14th international symposium on biomedical imaging (ISBI 2017), IEEE, 2017, pp. 986--990.

\bibitem{yu2018melanoma}
Z.~Yu, X.~Jiang, F.~Zhou, J.~Qin, D.~Ni, S.~Chen, B.~Lei, T.~Wang, Melanoma recognition in dermoscopy images via aggregated deep convolutional features, IEEE Transactions on Biomedical Engineering 66~(4) (2018) 1006--1016.

\bibitem{ozkan2017skin}
I.~A. OZKAN, M.~KOKLU, Skin lesion classification using machine learning algorithms, International Journal of Intelligent Systems and Applications in Engineering 5~(4) (2017) 285--289.

\bibitem{moradi2019kernel}
N.~Moradi, N.~Mahdavi-Amiri, Kernel sparse representation based model for skin lesions segmentation and classification, Computer methods and programs in biomedicine 182 (2019) 105038.

\bibitem{al2020automatic}
Z.~Al~Nazi, T.~A. Abir, Automatic skin lesion segmentation and melanoma detection: Transfer learning approach with u-net and dcnn-svm, in: Proceedings of International Joint Conference on Computational Intelligence, Springer, 2020, pp. 371--381.

\bibitem{rodrigues2020new}
D.~D.~A. Rodrigues, R.~F. Ivo, S.~C. Satapathy, S.~Wang, J.~Hemanth, P.~P. Reboucas~Filho, A new approach for classification skin lesion based on transfer learning, deep learning, and iot system, Pattern Recognition Letters 136 (2020) 8--15.

\bibitem{afza2021hierarchical}
F.~Afza, M.~Sharif, M.~Mittal, M.~A. Khan, D.~J. Hemanth, A hierarchical three-step superpixels and deep learning framework for skin lesion classification, Methods (2021).

\bibitem{habibzadeh2013white}
M.~Habibzadeh, A.~Krzy{\.z}ak, T.~Fevens, White blood cell differential counts using convolutional neural networks for low resolution images, in: International Conference on Artificial Intelligence and Soft Computing, Springer, 2013, pp. 263--274.

\bibitem{zhao2017automatic}
J.~Zhao, M.~Zhang, Z.~Zhou, J.~Chu, F.~Cao, Automatic detection and classification of leukocytes using convolutional neural networks, Medical \& biological engineering \& computing 55~(8) (2017) 1287--1301.

\bibitem{sharma2019white}
M.~Sharma, A.~Bhave, R.~R. Janghel, White blood cell classification using convolutional neural network, in: Soft Computing and Signal Processing, Springer, 2019, pp. 135--143.

\bibitem{derrac2011practical}
J.~Derrac, S.~Garc{\'\i}a, D.~Molina, F.~Herrera, A practical tutorial on the use of nonparametric statistical tests as a methodology for comparing evolutionary and swarm intelligence algorithms, Swarm and Evolutionary Computation 1~(1) (2011) 3--18.

\bibitem{oyewola2022novel}
D.~O. Oyewola, E.~G. Dada, S.~Misra, R.~Dama{\v{s}}evi{\v{c}}ius, A novel data augmentation convolutional neural network for detecting malaria parasite in blood smear images, Applied Artificial Intelligence (2022) 1--22.

\end{thebibliography}


\end{document}